%% file: main.tex
\RequirePackage{fix-cm}
\documentclass[twocolumn]{svjour3}           %
\smartqed  %
\usepackage{graphicx}
\usepackage{url}              %
\usepackage{booktabs}         %
\usepackage{amsfonts}         %
\usepackage{color}            %
\usepackage[table]{xcolor}    %
\usepackage{multirow}
\usepackage{adjustbox}
\usepackage{caption} %
\usepackage[misc]{ifsym}
\usepackage[pagebackref=true,breaklinks=true,letterpaper=true,colorlinks,bookmarks=false]{hyperref}

\def\sepappendix{0}

\newcommand{\cpfont}{\small}
\begin{document}

\title{Synthetic Humans for Action Recognition from Unseen Viewpoints
}

\author{G\"ul Varol$^{1~\textnormal{\scriptsize{\Letter}}}$ \and Ivan Laptev$^2$ \and Cordelia Schmid$^2$ \and Andrew Zisserman$^3$}
\authorrunning{Varol et al.} %

\institute{
    \Letter{} gul.varol@enpc.fr \\ \\
    $^1$LIGM, \'{E}cole des Ponts, Univ Gustave Eiffel, CNRS, France \\
    $^2$Inria, France \\
    $^3$Visual Geometry Group, University of Oxford, UK
}

\date{Received: 14 September 2020 / Accepted: 5 April 2021}

\maketitle

\input{abstract}
\input{introduction}
\input{relatedwork}
\input{approach}
\input{experiments}

\input{conclusions}

\begin{acknowledgements}
    This work was supported in part by Google Research,
    EPSRC grant ExTol,
    Louis Vuitton ENS Chair on Artificial Intelligence,
    DGA project DRAAF,
    and the French government under management of Agence Nationale de la Recherche as part of the Investissements d'Avenir program, reference ANR-19-P3IA-0001 (PRAIRIE 3IA Institute).
    We thank Angjoo Kanazawa,
    Fabien Baradel, and Max Bain for helpful discussions,
    Philippe Weinzaepfel and Nieves Crasto for providing pre-trained models.
\end{acknowledgements}

\bibliographystyle{splncs04}
\bibliography{references}

\bigskip
{\noindent \large \bf {APPENDIX}}\\
\input{appendix}

\end{document}

%% file: abstract.tex
\begin{abstract}
Although synthetic training data has been shown to be beneficial for tasks such as human pose estimation, its use for RGB human action recognition is relatively unexplored. Our goal in this work is to answer the question \textit{whether synthetic humans can improve the performance of human action recognition}, with a particular focus on generalization to unseen viewpoints. We make use of the recent advances in monocular 3D human body reconstruction from real action sequences to automatically render synthetic training videos for the action labels. We make the following contributions: (i) we investigate the extent of variations and augmentations that are beneficial to improving performance at new viewpoints. We consider changes in body shape and clothing for individuals, as well as more action relevant augmentations such as non-uniform frame sampling, and interpolating between the motion of individuals performing the same action; (ii)  We introduce a new data generation methodology, {\em SURREACT}, that allows training of spatio-temporal CNNs for action classification; (iii) We substantially improve the state-of-the-art action recognition performance on the NTU RGB+D and UESTC standard human action multi-view benchmarks; Finally, (iv) we extend the augmentation approach to in-the-wild videos from a subset of the Kinetics dataset to investigate the case when only one-shot training data is available, and demonstrate improvements in this case as well.

\keywords{Synthetic humans \and Action recognition}
\end{abstract}

%% file: introduction.tex
\section{Introduction}
\label{sec:introduction}

\begin{figure}
	\centering
	\includegraphics[width=.99\linewidth]{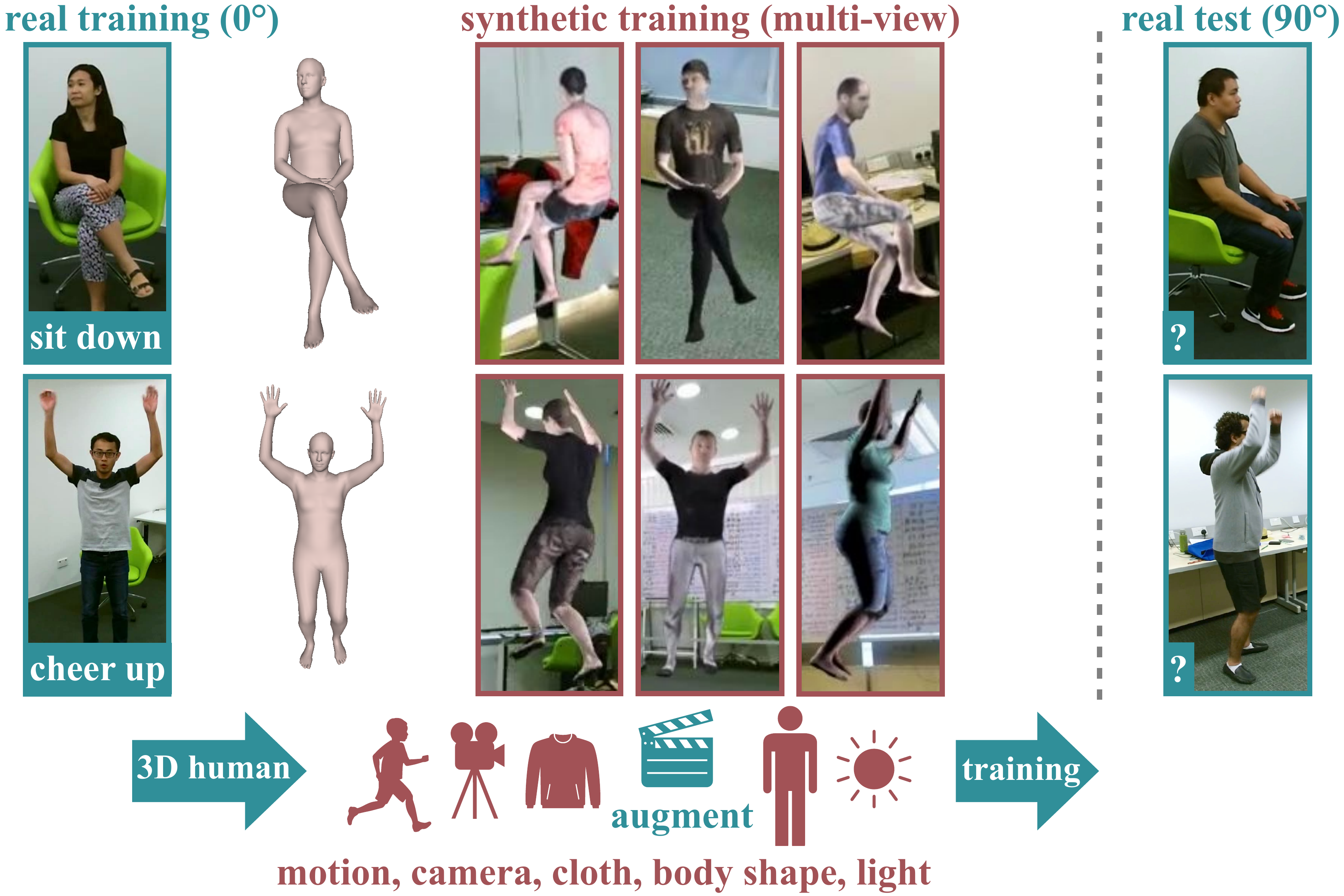}
    \vspace{-.6cm}
	\caption{
        \cpfont
        \textbf{Synthetic humans for actions:} We estimate 3D shape from real videos and automatically
		render synthetic videos with action labels. We explore
		various augmentations for motions, viewpoints, and appearance.
		Training temporal CNNs with this data significantly improves the action
		recognition from unseen viewpoints.
	}
	\label{fig:teaser}
    \vspace{-.4cm}
\end{figure}

Learning human action representations from RGB video data
has been widely studied.
Recent advances on convolutional
neural networks (CNNs)~\cite{LeCun1989cnn}
have shown excellent performance~\cite{Carreira2017,feichtenhofer2019slowfast,Feichtenhofer16,HaraCVPR2018,lin2019tsm,varol18_ltc,TSN2016ECCV}
on benchmark datasets, such as UCF101~\cite{UCF101}.
However, the success of CNNs rely heavily on
the availability of large-scale training data, which is not always the case.
To address the lack of training data, several works explore the use of
complementary synthetic data
for a range of tasks in computer vision such as
 optical flow estimation,
 segmentation,  human body and hand pose estimation~\cite{dosovitskiy_flownet,shotton2011,Su_2015_ICCV,varol2017surreal,zb2017hand}.
In this work, we raise the question \textit{how to synthesize
videos for action recognition} in the case of limited real data, such
as only one viewpoint, or one-shot available at training.

Imagine a surveillance or ambient assisted living system,
where a dataset is already collected for a set of actions from a certain camera.
Placing a new camera in the environment from a new viewpoint would require
re-annotating data because
the appearance of an
action is drastically different when performed from
different viewpoints~\cite{junejo2011,Liu2011CrossviewAR,Zheng2016}.
In fact, we observe that  state-of-the-art action recognition networks
fail drastically when trained and tested on distinct viewpoints.
Specifically, we train the model of~\cite{HaraCVPR2018} on videos from
a benchmark dataset NTU RGB+D~\cite{NTURGBD}
where people are facing the camera.
When we test this network
on other front-view (0$^\circ$) videos, we obtain $\sim$80\% accuracy. When we
test with side-view (90$^\circ$) videos, the performance drops
to $\sim$40\% (see Section~\ref{sec:experiments}).
This motivates us to study action recognition from novel viewpoints.

Existing methods addressing cross-view action recognition
do not work in challenging setups
(e.g.~same subjects and similar viewpoints in training and test splits~\cite{NTURGBD}).
We introduce and study a more challenging protocol
with
only one viewpoint
at training.
Recent methods assuming multi-view training data~\cite{NIPS2018_7401,Li2018,Wang_2018_ECCV} also become inapplicable.

A naive way to achieve generalization 
is to collect data from all views, for all possible conditions,
but this is impractical due to combinatorial explosion \cite{Yuille2018DeepNW}. 
Instead, we augment the existing real data synthetically
to increase the diversity in terms of viewpoints, appearance, and motions.
Synthetic humans are relatively easy to render
for tasks such as pose estimation, because arbitrary
motion capture (MoCap) resource can be used~\cite{shotton2011,varol2017surreal}.
However, action classification
requires certain motion patterns and semantics.
It is challenging to generate synthetic data with action labels~\cite{DeSouza:Procedural:CVPR2017}.
Typical MoCap datasets~\cite{cmu_mocap}, targeted for pose diversity, are not suitable for
action recognition due to lack of clean action annotations. Even if one collects a 
MoCap dataset, it is still limited to pre-defined set of
categories.

In this work, we propose
a new, efficient and scalable approach for generating
\textit{synthetic videos with action labels} from the target set of categories.
We employ a 3D human motion estimation
method, such as HMMR~\cite{humanMotionKZFM19} and VIBE~\cite{VIBECVPR2020},
that automatically extracts
the 3D human dynamics from a single-view RGB video.
The resulting sequence of SMPL body~\cite{smpl2015} pose parameters
are then combined with other randomized generation components
(e.g.~viewpoint, clothing) 
to render diverse complementary training data with action annotations.
Figure~\ref{fig:teaser} presents an overview of our pipeline.
We demonstrate the advantages of such data
when training spatio-temporal CNN models
for (i) action recognition from unseen viewpoints and
(ii) training with one-shot real data.
We boost performance on unseen viewpoints from 53.6\% to 69.0\% on NTU,
and from 49.4\% to 66.4\%
on UESTC dataset
by augmenting limited real training data with our proposed SURREACT dataset.
Furthermore, we present an in-depth analysis
about the importance of
action relevant augmentations
such as diversity of motions and viewpoints, as well as
our non-uniform frame sampling strategy which substantially
improves the action recognition performance.
Our code and data will be available at
the project page\footnote{\url{https://www.di.ens.fr/willow/research/surreact/}}.

%% file: relatedwork.tex
\section{Related Work}
\label{sec:relatedwork}

Human action recognition is a well-established research field.
For a broad review of the literature on action recognition,
see the recent survey of Kong~et~al.~\cite{Kong2018HumanAR}.
Here, we focus on relevant works on synthetic data,
cross-view action recognition, and briefly on 3D human
shape estimation.

\vspace{0.1cm}
\noindent\textbf{Synthetic humans.}
Simulating human motion dates back to 1980s.
Badler~et~al.~\cite{Badler1993} provide
an extensive overview of early approaches.
More recently, synthetic images of people
have been used to train visual models for
2D/3D body pose and shape estimation~\cite{synthetic_cohenor,Deep3DPose,Liu2019Temp,Pishchulin2012,shotton2011,varol18_bodynet},
part segmentation~\cite{shotton2011,varol2017surreal},
depth estimation~\cite{varol2017surreal},
multi-person pose estimation~\cite{Hoffmann:GCPR:2019},
pedestrian detection~\cite{MarinVGL10,Pishchulin2012},
person re-identification~\cite{Qian_2018_ECCV},
hand pose estimation~\cite{hasson19_obman,zb2017hand}, and
face recognition~\cite{Kortylewski2018,Masi2019Face}.
Synthetic datasets built for these tasks, such as the recent
SURREAL dataset \cite{varol2017surreal}, however,
do not provide action labels.

Among previous works that focus on synthetic
human data, %
very few tackle action recognition~\cite{DeSouza:Procedural:CVPR2017,Liu2019,Rahmani2016Novel}.
Synthetic 2D human pose sequences~\cite{Lv2007}
and synthetic point trajectories~\cite{NKTM,rahmani2018,Jingtian2018} have been used
for view-invariant action recognition.
However,
RGB-based synthetic training for action recognition is relatively new,
with \cite{DeSouza:Procedural:CVPR2017} being one of the first attempts.
De Souza~et~al. \cite{DeSouza:Procedural:CVPR2017} manually define 35 action classes
and jointly estimate real categories and synthetic categories in a multi-task setting.
However, their categories are not easily scalable and do not necessarily relate to the target set of classes.
Unlike~\cite{DeSouza:Procedural:CVPR2017},
we automatically extract motion sequences from real data,
making the method flexible for new categories.
Recently, \cite{virtualhome2018} has generated the VirtualHome dataset,
a simulation environment with
programmatically defined synthetic activities 
using crowd-sourcing. Different than our work,
the focus of~\cite{virtualhome2018} is not generalization to real data.

Most relevant to ours, \cite{Liu2019} generates synthetic training images
to achieve better performance on unseen viewpoints.
The work of~Liu~et~al.~\cite{Liu2019} is an extension of~\cite{Rahmani2016Novel}
by using RGB-D as input instead of depth only.
Both works formulate a frame-based pose classification
problem on their synthetic data, which they then use as features for action recognition.
These features
are not necessarily discriminative for the target action categories.
Different than this direction, %
we explicitly assign an action
label to synthetic videos and define the supervision directly
on action classification.

\vspace{0.1cm}
\noindent\textbf{Cross-view action recognition.}
Due to the difficulty of building multi-view action
recognition datasets, the standard benchmarks have been
recorded in controlled environments. RGB-D datasets
such as IXMAS~\cite{IXMAS2007}, UWA3D~II~\cite{Rahmani2014} and N-UCLA~\cite{NUCLA2014}
were state of the art %
until the availability of the large-scale NTU RGB+D dataset~\cite{NTURGBD}.
The size of NTU allows training deep neural
networks unlike previous datasets. %
Very recently, Ji~et al.~\cite{hri40} collected the first large-scale
dataset, UESTC, that has a 360$^\circ$ coverage around the performer, although
still in a lab setting.

Since multi-view action datasets are typically
captured with depth sensing devices, such as Kinect, they also
provide an accurate estimate of the 3D skeleton. Skeleton-based
cross-view action recognition therefore received a lot of attention
in the past decade~\cite{KeCVPR17,LiuECCV2016,LiuCVPR2017,Liu2017ESV,Zhang2017ViewAdaptive}.
Variants of LSTMs~\cite{LSTM1997} have been widely used~\cite{LiuECCV2016,LiuCVPR2017,NTURGBD}.
Recently, spatio-temporal skeletons were represented as images~\cite{KeCVPR17} or
higher dimensional objects~\cite{Liu2017ESV} where standard CNN
architectures were applied.

RGB-based cross-view action recognition is in comparison less studied.
Transforming RGB features to be view-invariant is not
as trivial as transforming 3D skeletons.
Early work on transferring appearance features from the source view
to the target view explored the use of maximum margin clustering
to build a joint codebook for temporally synchronous videos~\cite{Farhadi2008}.
Following this approach, several other works focused on building
global codebooks to extract view-invariant
representations~\cite{KongTIP2017,Liu2019_viewinvariant,rahmani2018,Zheng2013,Zheng2016}.
Recently, end-to-end approaches used human pose information as guidance for
action recognition~\cite{Baradel17,Liu_2018_CVPR,Luvizon20182D3DPE,ZolfaghariOSB17}.
Li~et~al.~\cite{NIPS2018_7401} formulated an adversarial view-classifier
to achieve view-invariance.
Wang~et~al.~\cite{Wang_2018_ECCV} proposed to fuse view-specific features
from a multi-branch CNN.
Such approaches cannot handle single-view training~\cite{NIPS2018_7401,Wang_2018_ECCV}.
Our method differs from these works
by compensating for
the lack of view diversity with synthetic videos.
We augment the real data
automatically at training time, and our model does not involve
any extra cost at test time unlike~\cite{Wang_2018_ECCV}.
Moreover, we do not assume real multi-view videos at training.

\vspace{0.1cm}
\noindent\textbf{3D human shape estimation.}
Recovering the full human body mesh from a single image
has been explored as a model-fitting problem~\cite{Bogo2016smplify,lassner2017up},
as regressing model parameters with CNNs~\cite{hmrKanazawa17,omran2018nbf,pavlakos2018humanshape,tung2017selfsupervised},
and as regressing non-parametric representations such as graphs or volumes~\cite{kolotouros2019cmr,varol18_bodynet}.
Recently, CNN-based parameter regression approaches have been
extended to video~\cite{humanMotionKZFM19,Liu2019Temp,VIBECVPR2020}.
HMMR~\cite{humanMotionKZFM19} builds on the single-image-based HMR~\cite{hmrKanazawa17}
to learn the human dynamics by using 1D temporal convolutions.
More recently, VIBE~\cite{VIBECVPR2020} adopts a recurrent model
based on frame-level pose estimates provided by SPIN~\cite{kolotouros2019spin}.
VIBE also incorporates an adversarial loss that
penalizes the estimated pose sequence if it is not a `realistic' motion, i.e., indistinguishable
from the real AMASS~\cite{AMASS:ICCV:2019} MoCap sequences.
In this work,
we recover 3D body parameters from real videos using
HMMR~\cite{humanMotionKZFM19} and VIBE~\cite{VIBECVPR2020}.
Both methods employ the SMPL body model~\cite{smpl2015}.
We provide a comparison between the two methods
for our purpose of action recognition, which can serve
as a proxy task to evaluate motion estimation.

%% file: approach.tex
\section{Synthetic humans with action labels}
\label{sec:approach}
Our goal is to improve the performance of action recognition using synthetic data
in cases where the real data is limited, e.g. domain mismatch between training/test
such as viewpoints or low-data regime.
In the following, we describe the three stages of: (i)~obtaining 3D temporal models for human actions
from real training sequences (at a particular viewpoint) (Section~\ref{subsec:approach:estimation}); (ii)~using these 3D temporal models to generate training sequences for new (and the original) viewpoints using
a rendering pipeline with augmentation 
(Section~\ref{subsec:approach:surreact}); and (iii)~training 
a spatio-temporal CNN with both real and synthetic data
(Section~\ref{subsec:approach:training}).

\subsection{3D human motion estimation}
\label{subsec:approach:estimation}

In order to generate a synthetic video with graphics techniques, we need to have
a sequence of articulated 3D human body models.
We employ the parametric body model SMPL~\cite{smpl2015},
which is a statistical model, learned over thousands of 3D scans.
SMPL generates the mesh of a person given
the disentangled pose and shape parameters.
The pose parameters ($\mathbb{R}^{72}$) control the kinematic deformations
due to skeletal posture, while the shape parameters ($\mathbb{R}^{10}$)
control identity-specific deformations such as the person height.

We hypothesize that a human action can be captured by
the sequence of \textit{pose} parameters,
and that the \textit{shape} parameters are largely irrelevant
(note, this may not necessarily be true for human-object interaction categories).
Given reliable 3D pose
sequences from
action recognition video datasets, 
we can transfer the associated action labels to synthetic videos.
We use the recent method of
Kanazawa~et~al.~\cite{humanMotionKZFM19},
namely human mesh and motion recovery (HMMR), unless
stated otherwise.
HMMR extends the single-image reconstruction method
HMR~\cite{hmrKanazawa17} to video with a multi-frame CNN
that takes into account a temporal neighborhood around a video frame.
HMMR learns a temporal representation
for human dynamics by incorporating large-scale
2D pseudo-ground truth poses for in-the-wild videos.
It uses PoseFlow~\cite{Zhang_2018_CVPR} and AlphaPose~\cite{fang2017rmpe}
for multi-person 2D pose estimation and tracking as a pre-processing step.
Each person crop is then given as input
to the CNN for estimating the pose and shape, as well as
the weak-perspective camera parameters.
We refer the
reader to~\cite{humanMotionKZFM19} for more details.
We choose this method for the robustness on in-the-wild
videos, ability to capture multiple people, and the smoothness of the recovered motion,
which are important for our generalization from synthetic videos
to real. Figure~\ref{fig:teaser} presents the 3D pose
animated synthetically for sample video frames.
We also experiment with the more recent motion estimation method, VIBE~\cite{VIBECVPR2020},
and show that improvements in motion estimation proportionally affect
the action recognition performance in our pipeline.
Note that
we only use the pose parameters from HMMR or VIBE, and randomly
change the shape parameters, camera parameters, and other factors. Next, we
present the augmentations in our synthetic data generation.

\subsection{SURREACT dataset components}
\label{subsec:approach:surreact}
\begin{figure}[t]
    \centering
    \includegraphics[width=.99\linewidth]{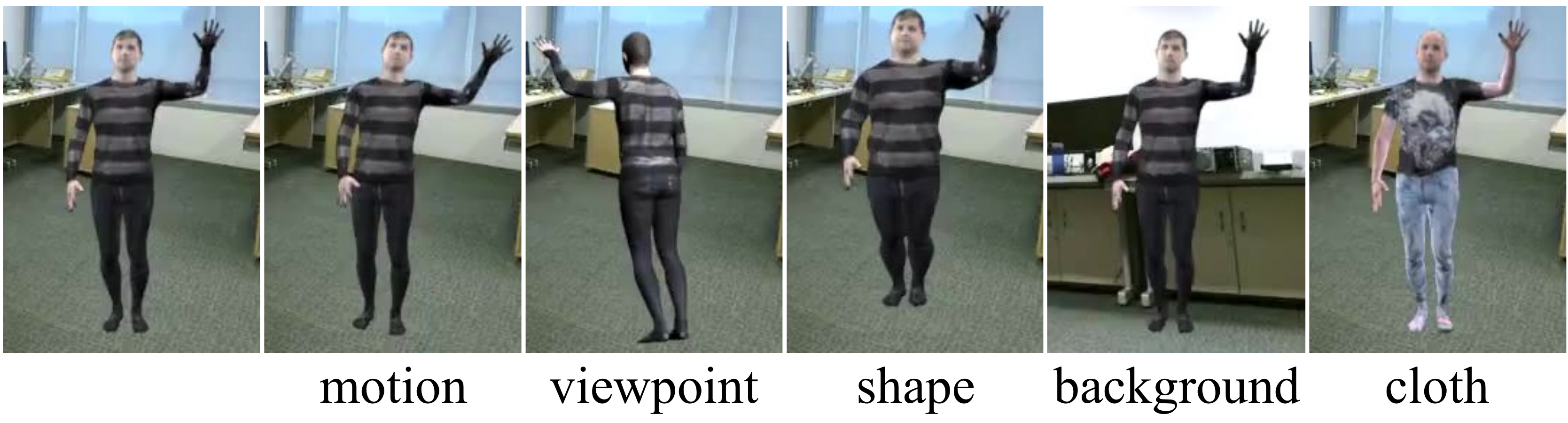}
    \caption{
        \cpfont 
        \textbf{Augmentations:}
        We illustrate different augmentations of the SURREACT dataset
        for the hand waving action.
        We modify the joint angles with additive noise on the pose parameters for \textit{motion}
        augmentation. We systematically change the camera position to create \textit{viewpoint}
        diversity. We sample from a large set of body \textit{shape} parameters, \textit{backgrounds},
        and \textit{clothing} to randomize appearances.
    }
    \label{fig:augmentations}
\end{figure}

\begin{figure*}
    \includegraphics[width=0.99\linewidth]{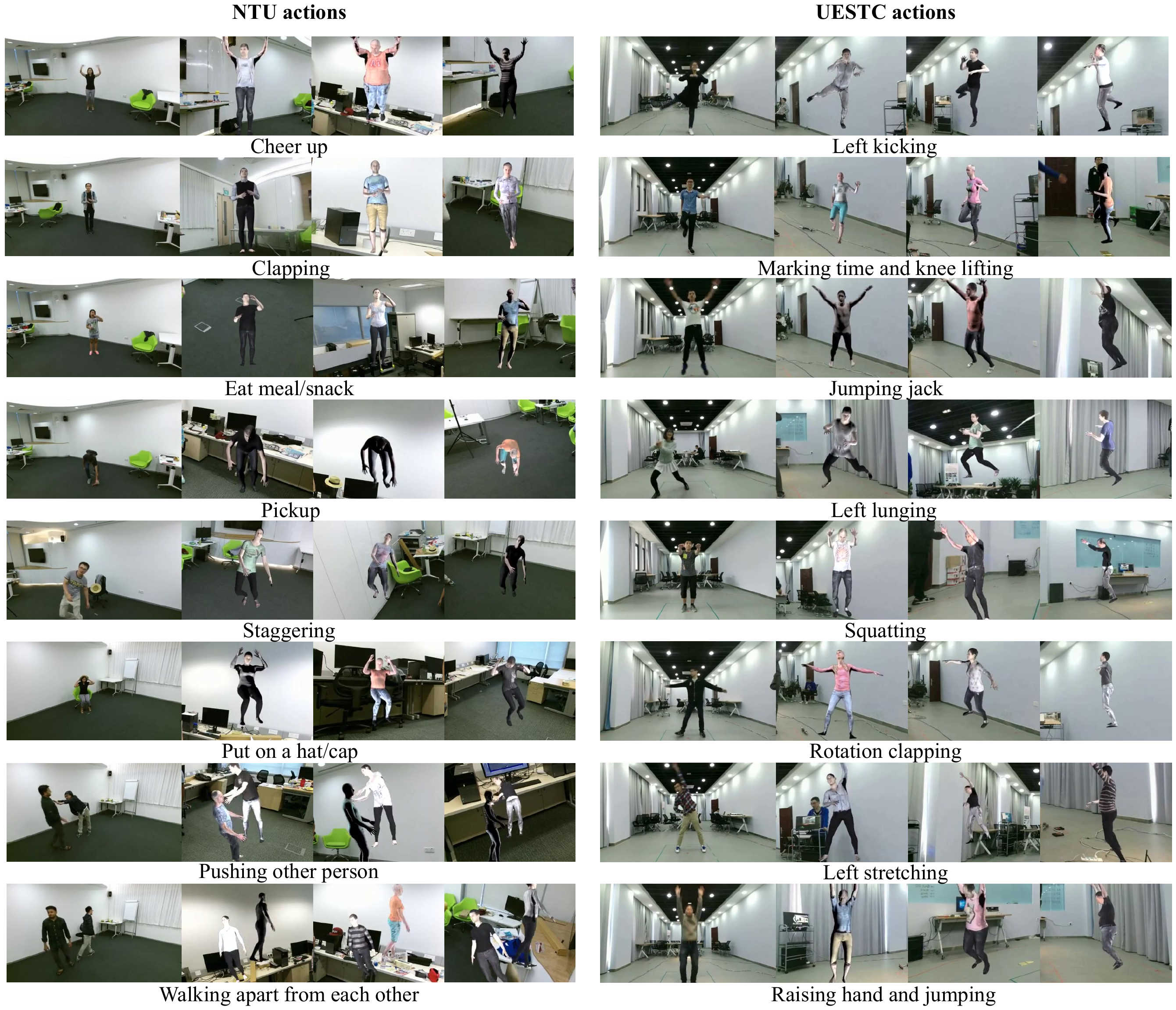}
    \vspace{-0.3cm}
    \caption{
        \cpfont 
        \textbf{SURREACT:} We visualize samples from SURREACT for the actions from the NTU (left)
        and the UESTC (right) datasets. The motions are estimated using HMMR.
        Each real video frame is accompanied with three synthetic
        augmentations. On the left, we show the variations in clothes, body shapes, backgrounds,
        camera height/distance
        from the original 0$^\circ$ viewpoint. On the right, we show the variations in viewpoints
        for 0$^\circ$, 45$^\circ$, and 90$^\circ$ views. The complete list of actions can be found
        as a video at the project page~\cite{surreactpage}.
    }
    \vspace{-0.5cm}
    \label{fig:surreact}
\end{figure*}

In this section, we give details on our synthetic dataset,
SURREACT (Synthetic hUmans foR REal ACTions).
We follow \cite{varol2017surreal}
and render 3D SMPL sequences with
randomized cloth textures, lighting, and body shapes.
We animate the body model with our automatically extracted
pose dynamics as described in the previous section.
We explore various \textit{motion augmentation}
techniques to increase intra-class diversity
in our training videos.
We incorporate \textit{multi-person} videos which
are especially important for two-people
interaction categories.
We also systematically sample from 8 \textit{viewpoints} around a circle
to perform controlled experiments.
Different augmentations are illustrated in Figure~\ref{fig:augmentations}
for a sample synthetic frame.
Visualizations from SURREACT are further provided in Figure~\ref{fig:surreact}.

Each generated video has automatic ground truth for 3D joint locations,
part segmentation, optical flow, and SMPL body \cite{smpl2015} parameters,
as well as an action label, which we use for training
a video-based 3D CNN for action classification.
We use other ground truth modalities as input to
action recognition as oracle experiments
\if\sepappendix1{(see the supplemental material).}
\else{(see Table~\ref{table:app:synthinputs}).}
\fi
We further use the optical flow ground truth to train a flow estimator
and use the segmentation to randomly augment the background
pixels in some experiments.

Our new SURREACT dataset differs from the SURREAL dataset \cite{varol2017surreal}
mainly by providing action labels, exploring motion augmentation, and by using automatically extracted
motion sequences instead of MoCap recordings~\cite{cmu_mocap}.
Moreover, Varol~et~al.~\cite{varol2017surreal} do not exploit the temporal
aspect of their dataset, but only train CNNs with single-image input.
We further employ multi-person videos and a systematic viewpoint distribution.

\vspace{0.1cm}
\noindent\textbf{Motion augmentation.}
Automatic extraction of 3D sequences
from 2D videos poses an additional challenge
in our dataset compared to clean high-quality
MoCap sequences. To reduce the jitter, %
we temporally smooth
the estimated SMPL pose parameters
by weighted linear averaging.
SMPL poses are represented as axis-angle rotations
between joints. We convert them into quaternions 
when we apply linear operations, then normalize each quaternion
to have a unit norm, before converting back to axis-angles.
Even with this processing, the motions may remain
noisy, which is inevitable given that monocular
3D motion estimation is a difficult task on its own.
Our findings interestingly suggest that the synthetic
human videos are still beneficial when the motions are noisy.

To increase motion diversity,
we further perturb the pose parameters
with various augmentations. Specifically,
we use a video-level \textit{additive noise} on the
quaternions for each body joint to slightly change the poses,
as an intra-individual augmentation.
We also experiment with an inter-individual augmentation
by interpolating
between motion
sequences of the same action class.
Given a pair of sequences from two individuals, we first align them
with dynamic time warping~\cite{Sakoe1978DynamicPA}, %
then we linearly
interpolate the quaternions of the time-aligned sequences to generate
a new sequence, which we refer as \textit{interpolation}.
A visual explanation of the process can be found in
\if\sepappendix1{the supplemental material.}
\else{Figure~\ref{fig:app:interpolation}.}
\fi
We show significant gains by increasing motion diversity.

\vspace{0.1cm}
\noindent\textbf{Multi-person.}
We use the 2D pose information from~\cite{fang2017rmpe,Zhang_2018_CVPR}
to count the number of people in the real video.
In the case of a single-person,
we center the person on the image and do not add 3D translation to the body,
i.e., the person is centered independently for each frame.
While such constant global positioning of the body loses information
for some actions such as \textit{walking} and \textit{jumping},
we find that the translation estimate
adds more noise to consider this information
and potentially increases the domain gap with the real where
no such noise exists
(see
\if\sepappendix1{the supplemental material).}
\else{Appendix~\ref{sec:app:implementation}).}
\fi
If there is more than one person,
we insert additional body model(s) for rendering. 
We translate each person
in the $xy$ image plane. Note that we do not translate the person in full $xyz$ space.
We observe that the $z$ component of the translation
estimation is not reliable due to the depth ambiguity therefore the people
are always centered at $z=0$. More explanations about
the reason for omitting the $z$ component can be found in
\if\sepappendix1{the supplemental material.}
\else{Appendix~\ref{sec:app:implementation}.}
\fi
We temporally smooth the translations to reduce the noise.
We subtract the mean of
translations across the video and across the people to roughly
center all people to the frame. We therefore keep the relative distances between people,
which is important for actions such as \textit{walking towards each other}.

\vspace{0.1cm}
\noindent\textbf{Viewpoints.}
We systematically render each motion sequence 8 times by
randomizing all other generation parameters at each view.
In particular, we place the camera to be rotated at
\{$0^{\circ}, 45^{\circ}, 90^{\circ}, 135^{\circ}, 180^{\circ}, 225^{\circ}, 270^{\circ},$
 $315^{\circ}$\}
azimuth angles with respect to the origin, denoted as
($0^\circ$:$45^\circ$:$360^\circ$) in our experiments.
The distance of the camera from the origin and the height of the camera
from the ground are randomly sampled from a predefined range: $[4, 6]$ meters for the distance,
$[-1, 3]$ meters for the height.
This can be adjusted according to the target test setting.

\vspace{0.1cm}
\noindent\textbf{Backgrounds.} Since we have access to the target
real dataset where we run pose estimation methods, we can extract
background pixels directly from the training set of this dataset.
We crop from regions without the person to obtain static
backgrounds for the NTU and UESTC datasets. We experimentally
show the benefits of using the target dataset backgrounds in the Appendix
\if\sepappendix1{(see Table A.5).}
\else{(see Table~\ref{table:app:backgrounds}).}
\fi
For Kinetics
experiments, we render human bodies on top of
unconstrained videos from non-overlapping
action classes and show benefits over static backgrounds.
Note that these background videos might also include
human pixels.

\subsection{Training 3D CNNs with non-uniform frames}
\label{subsec:approach:training}

Following the success of 3D CNNs for video
recognition~\cite{Carreira2017,HaraCVPR2018,C3D}, %
we employ a spatio-temporal convolutional
architecture that operates on multi-frame video inputs.
Unless otherwise specified, our network architecture is 3D ResNet-50~\cite{HaraCVPR2018}
and its weights are randomly initialized
(see Appendix~\ref{subsec:app:pretraining} for pretraining experiments).

To study the generalization capability of synthetic data
across different input modalities,
we train one CNN for RGB and another for optical flow
as in \cite{simonyan_twostream}. We average the scores with
equal weights
when reporting the fusion.

We subsample fixed-sized inputs from videos to have a
$16\times256\times 256$ spatio-temporal resolution,
in terms of number of frames, width, and height, respectively.
In case of optical flow input, we map the RGB input to
$15\times64\times 64$ dimensional flow estimates. To estimate flow, we train a two-stack hourglass
architecture \cite{newell2016hourglass} with our synthetic data for flow estimation
on 2 consecutive frames.
We refer the reader to
\if\sepappendix1{the supplemental material}
\else{Figure~\ref{fig:app:flowntu}}
\fi
for the qualitative results of our optical flow estimation.

\begin{figure}
    \centering
    \includegraphics[width=0.99\linewidth]{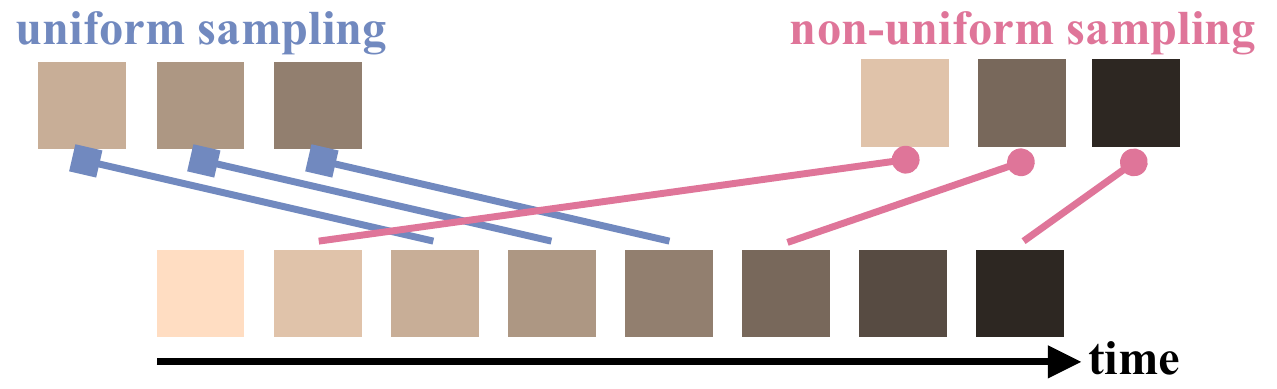}
    \mbox{}\vspace{-0.6cm}
    \caption{
        \cpfont 
        \textbf{Frame sampling:} We illustrate our non-uniform frame sampling strategy
        in our 3D CNN training. Compared to the commonly adopted consecutive setting
        which uniformly samples with a fixed frame rate, non-uniform
        sampling has random skips in time, allowing speed augmentations
        and long-term context.}
    \mbox{}\vspace{-.8cm}
    \label{fig:nonuniform}
\end{figure}

\vspace{0.1cm}
\noindent\textbf{Non-uniform frame sampling.}
We adopt a different frame sampling strategy than most works~\cite{Carreira2017,feichtenhofer2019slowfast,HaraCVPR2018}
in the context of 3D CNNs.
Instead of uniformly sampling (at a fixed frame rate) a video clip with consecutive frames,
we randomly sample frames across time by keeping
their temporal order, which we refer as
\textit{non-uniform} sampling.
Although recent works explore multiple temporal resolutions, e.g.~by
regularly sampling at two different frame rates~\cite{feichtenhofer2019slowfast},
or randomly selecting a frame rate~\cite{Zhu2018RandomTS},
the sampled frames are equidistant from each other.
TSN~\cite{TSN2016ECCV} and ECO~\cite{ECO_eccv18} employ a hybrid strategy by regularly sampling
temporal segments and randomly sampling a frame from each segment,
which is a more restricted special case of our strategy.
Moreover, TSN uses a 2D CNN without temporal modelling.
\cite{ECO_eccv18} also has 2D convolutional features on each frame,
which are stacked as input to a 3D CNN only at the end of the network. %
None of these works provide controlled experiments to
quantify the effect of their sampling strategy.
The concurrent work of \cite{chen2020deep} presents
an experimental analysis
comparing the dense consecutive sampling with the hybrid sampling
of TSN.

Figure~\ref{fig:nonuniform} compares the consecutive sampling
with our non-uniform sampling.
In our experiments,
we report results for both and show improvements for the latter.
Our videos are temporally trimmed around the action, therefore,
each video is short, i.e.~spans several seconds.
During training we randomly sample 16 video frames
as a fixed-sized input to 3D CNN.
Thus, the convolutional kernels become speed-invariant to some degree.
This can be seen as a data augmentation technique, as well as a way to
capture long-term cues.

\begin{figure*}[t]
    \centering
    \includegraphics[width=0.99\linewidth]{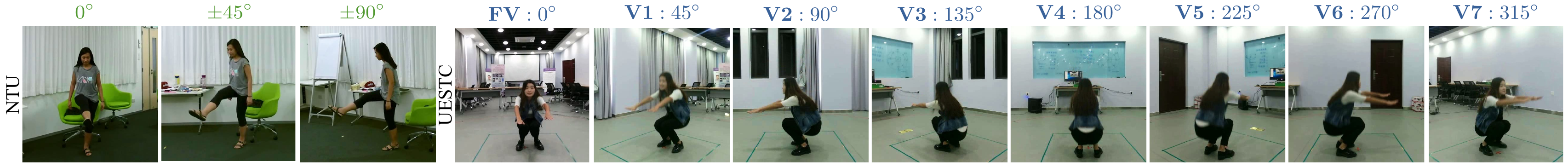}
    \caption{
        \cpfont 
        \textbf{Datasets:} We show sample video frames from the multi-view datasets used in our experiments.
        NTU and UESTC datasets have 3 and 8 viewpoints, respectively.
        NTU views correspond to $0^\circ$, $45^\circ$, and $90^\circ$ from
        left to right. UESTC
        covers $360^\circ$ around the performer.}
    \label{fig:datasets}
\end{figure*}

At test time, we sample several 16-frame clips and average the softmax scores.
If we test the uniform case, we sample non-overlapping
consecutive clips with sliding window. For the non-uniform case,
we randomly sample as many non-uniform clips as the number of sliding windows for the uniform case.
In other words, the number of sampled clips is proportional to the video length.
More precisely, let $T$ be the number of frames in the entire test video, $F$ be the number of input frames per clip, and S be the stride parameter. We sample $N$ clips where $N = \left\lceil(T - F) / S)\right\rceil + 1$. In our case $F=16$, $S=16$. We apply sliding window for the uniform case. For the non-uniform case, we sample $N$ clips, where each clip is an ordered random (without replacement) 16-frame subset from $T$.
We observe that it is important to train and test with the same sampling scheme,
and keeping the temporal order is important.
More details can be found in
\if\sepappendix1{the supplemental material.}
\else{Appendix~\ref{subsec:app:nonuniform}.}
\fi

\vspace{0.1cm}
\noindent\textbf{Synth+Real.} Since each real video
is augmented multiple times (e.g.~8 times for 8 views),
we have more synthetic data than real.
When we add synthetic data to training, we balance the real
and synthetic datasets such that at each epoch we randomly
subsample from the synthetic videos to have equal number
for both real and synthetic.

We minimize the cross-entropy loss using
RMSprop~\cite{Tieleman2012} with mini-batches of size 10
and an initial learning rate of $10^{-3}$ with a fixed schedule.
Color augmentation is used for the RGB stream. %
Other implementation details are given in
\if\sepappendix1{the supplemental material.}
\else{Appendix~\ref{sec:app:implementation}.}
\fi

%% file: experiments.tex
\section{Experiments}
\label{sec:experiments}

In this section, we start by presenting the action recognition
datasets
used in our experiments (Section~\ref{subsec:experiments:datasets}).
Next, we present extensive ablations for action recognition from unseen viewpoints (Section~\ref{subsec:experiments:ablation}).
Then, we compare our results to the state of the art for completeness (Section~\ref{subsec:experiments:sota}).
Finally, we illustrate our approach on in-the-wild videos (Section~\ref{subsec:experiments:oneshot}).

\subsection{Datasets and evaluation protocols}
\label{subsec:experiments:datasets}

We briefly present the datasets
used in this work,
as well as the evaluation protocols employed.

\vspace{0.1cm}
\noindent\textbf{NTU RGB+D dataset (NTU).} This dataset~\cite{NTURGBD} captures 60 actions with 3
synchronous cameras (see Figure~\ref{fig:datasets}).
The large scale (56K videos) of the dataset allows training deep neural networks.
Each sequence has 84 frames on average.
The standard protocols~\cite{NTURGBD} report accuracy for cross-view and cross-subject splits.
The cross-view (CV)
split considers 0$^\circ$ and 90  views as training and 45$^\circ$ view as test, and
the same subjects appear both in training and test.
For the cross-subject (CS) setting, 20 subjects are used for training, the remaining 20 for test, and
all 3 views are seen at both training and test.
We report on the standard protocols to be
able to compare to the state of the art (see Table~\ref{table:ntusota}).
However, we introduce a new protocol to
make the task more challenging.
From the cross-subject training split that has all 3 views,
we take only 0$^\circ$ viewpoint for training,
and we test on the 0$^\circ$, 45$^\circ$, 90$^\circ$ views of the cross-subject test split. 
We call this protocol cross-view-subject (CVS).
Our focus is mainly to improve for the unseen and distinct view of 90$^\circ$.

\vspace{0.1cm}
\noindent\textbf{UESTC RGB-D varying-view 3D action dataset (UESTC).} UESTC is a recent dataset~\cite{hri40}
that systematically collects 8 equally separated viewpoints
that cover 360$^\circ$ around a person (see Figure~\ref{fig:datasets}).
In total, the dataset has 118 subjects, 40 actions categories, and
26500 videos of more than 200 frames each.
This dataset allows studying actions from unusual views such as behind the person.
We use the official protocol Cross View I (CV-I), suitable for our task,
which trains with 1 viewpoint and tests with all other 7 for each view.
The final performance is evaluated as the average across all tests.
For completeness,
we also report the Cross View II (CV-II) protocol that concentrates on multi-view training,
i.e., training with even viewpoints (FV, V2, V4, V6)
and testing with odd viewpoints (V1, V3, V5, V7),
and vice versa.

\setlength{\tabcolsep}{8pt}
\begin{table*}
    \centering
    \resizebox{0.99\linewidth}{!}{
        \begin{tabular}{lccc | ccc | ccc }
            \toprule
            & \multicolumn{3}{c|}{RGB} & \multicolumn{3}{c|}{Flow} & \multicolumn{3}{c}{RGB + Flow}\\
            Training data & $0^\circ$ & \cellcolor{gray!25} $45^\circ$ & \cellcolor{gray!25} $90^\circ$  & $0^\circ$ & \cellcolor{gray!25} $45^\circ$  & \cellcolor{gray!25} $90^\circ$ & $0^\circ$ & \cellcolor{gray!25} $45^\circ$  & \cellcolor{gray!25} $90^\circ$ \\
            \midrule
            Real($0^\circ$) & 86.9 & 74.5 & 53.6 & 82.8 & 70.6 & 49.7 & 88.8 & 78.2 & 57.3 \\
            \midrule
            Synth$_\mathit{HMMR}$($0^\circ$:$45^\circ$:$360^\circ$) & 54.0 & 49.5 & 42.7 & 51.7 & 46.9 & 38.6 & 60.6 & 55.5 & 47.8 \\
            Synth$_\mathit{HMMR}$($0^\circ$:$45^\circ$:$360^\circ$) + Real($0^\circ$) & \textbf{89.1} & \textbf{82.0} & \textbf{67.1} & \textbf{85.9} & \textbf{76.4} & \textbf{58.9} & \textbf{90.5} & \textbf{83.3} & \textbf{68.0}\\
            \midrule
            Synth$_\mathit{VIBE}$($0^\circ$:$45^\circ$:$360^\circ$) & 58.1 & 52.8 & 45.3 & 54.1 & 47.2 & 37.9 & 63.0 & 57.6 & 48.3 \\
            Synth$_\mathit{VIBE}$($0^\circ$:$45^\circ$:$360^\circ$) + Real($0^\circ$) & \textbf{89.7} & \textbf{82.0} & \textbf{69.0} & \textbf{85.9} & \textbf{77.7} & \textbf{61.8} & \textbf{90.6} & \textbf{83.4} & \textbf{71.1} \\
            \bottomrule
        \end{tabular}
    }
    \mbox{}\vspace{-.1cm}
    \caption{
        \cpfont
        \textbf{Training jointly on synthetic and real data} substantially
        boosts the performance compared to only real training on NTU CVS protocol,
        especially on highlighted unseen views (e.g.,~69.0\% vs 53.6\%).
        The improvement can be seen for both RGB and Flow streams, as well as the fusion.
        We note the marginal improvements with the addition of flow unlike in other tasks
        where flow has been used to reduce the synthetic-real domain gap~\cite{Doersch2019}.
        We render two different versions of the synthetic dataset using HMMR
        and VIBE motion estimation methods, and observe improvements with VIBE.
        Moreover, training on synthetic videos alone is able to obtain 63.0\%
        accuracy.}
    \label{table:ntusynth}
    \mbox{}\vspace{-.6cm}
\end{table*}

 \begin{table}
    \centering
    \resizebox{0.99\linewidth}{!}{
        \begin{tabular}{llc @{\hspace{-0.1cm}}c @{\hspace{-0.1cm}}c | c @{\hspace{-0.3cm}} c @{\hspace{-0.3cm}}c}
            \toprule
            &                                       & \multicolumn{6}{@{\hspace{-0.3cm}}c}{Test Views} \\
            &                                       &               & uniform   &               &               & non-uniform   &               \\
            &                                       & $0^\circ$     & $45^\circ$    & $90^\circ$    & $0^\circ$     & $45^\circ$        & $90^\circ$    \\
            \midrule
            \multirow{5}{*}{\rotatebox[origin=c]{90}{Train Views}} & $0^\circ$                             & \textbf{83.9} & 67.9          & 42.9          & \textbf{86.9} & 74.5              & 53.6          \\
            & $45^\circ$                            & 72.1          & \textbf{81.6} & 66.8          & 78.1          & \textbf{85.2}     & 75.7          \\
            & $90^\circ$                            & 41.7          & 63.4          & \textbf{81.4} & 52.3          & 71.2              & \textbf{85.4} \\
            \cmidrule{2-8}
            & $0^\circ$ + $45^\circ$                & 86.0          & 85.3          & 69.9          & 89.7          & 88.9              & 79.3          \\
            & $0^\circ$ + $45^\circ$  + $90^\circ$  & \textbf{86.8} & \textbf{86.9} & \textbf{84.1} & \textbf{89.4} & \textbf{89.4}     & \textbf{87.8} \\
            \bottomrule
        \end{tabular}
    }
    \caption{
        \cpfont 
        \textbf{Real baselines:} Training and testing with our cross-view-subject (CVS) protocol of the NTU dataset using only real RGB videos.
        Rows and columns correspond to training and testing sets, respectively.
        Training and testing on the same viewpoint shows the best performance as can be seen by the diagonals of the first three rows.
        This shows the domain gap present between $0^\circ$, $45^\circ$, $90^\circ$ viewpoints.
        If we add more viewpoints to the training (last two rows) we account for the domain gap.
        Non-uniform frame sampling (right) consistently outperforms the uniform frame sampling (left).}
    \label{table:ntubaselines}
    \mbox{}\vspace{-.6cm}
\end{table}

\begin{figure*}
    \centering
    \includegraphics[width=0.89\linewidth]{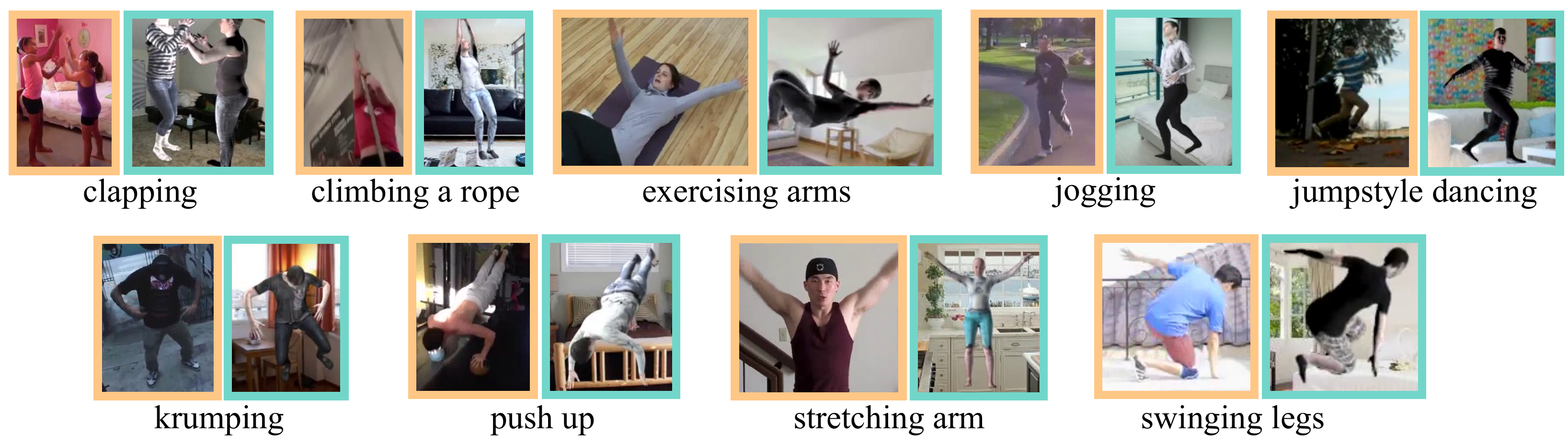}
    \caption{
        \cpfont 
        Sample video frames from the one-shot Kinetics-15 dataset.
        We provide side-by-side illustrations for real frames and their synthetically
        augmented versions from the original viewpoint. Note that we render the synthetic
        body on a static background for computational efficiency, but augment
        it during training with random real videos by using the segmentation mask.}
    \label{fig:kinetics}
\end{figure*}

\vspace{0.1cm}
\noindent\textbf{One-shot Kinetics-15 dataset (Kinetics-15).}
Since we wish to formulate a one-shot scenario from in-the-wild Kinetics \cite{Kinetics} videos,
we need a pre-trained model to serve as feature extractor.
We use a model pre-trained on Mini-Kinetics-200~\cite{Xie2017RethinkingSF},
a subset of Kinetics-400.
We define the novel classes from the remaining 200 categories
which can be described by body motions.
This procedure resulted in a 15-class subset of Kinetics-400:
\textit{bending back, clapping, climbing a rope, exercising arm, hugging,
jogging, jumpstyle dancing, krumping, push up, shaking hands, skipping rope,
stretching arm, swinging legs, sweeping floor, wrestling}.
Note that many of the categories such as \textit{waiting in line, dining, holding snake}
cannot be recognized solely by their body motions, but additional
contextual cues are needed.
From the 15 actions, we randomly sample 1 training video per class
(see Figure~\ref{fig:kinetics} for example videos with their synthetic
augmentations). The training set therefore consists of 15 videos.
For testing,
we report accuracy on all 725 validation videos from these 15 classes.
The limitation of this protocol is that it is sensitive to the choice
of the 15 training videos, e.g.,
if 3D motion estimation fails on one video,
the model will not benefit from additional synthetic data of one class.
Future work can consider multiple possible 
training sets (e.g., sampling videos where 3D pose estimation is confident) and 
report average performance.

\subsection{Ablation Study}
\label{subsec:experiments:ablation}
We first
compare real-only (Real),
synthetic-only (Synth), and mixed synthetic and real (Synth+Real)
training.
Next, we explore the effect of the motion estimation quality
and inputting raw motion parameters as opposed to synthetic renderings.
Then, we experiment
with the different synthetic data generation parameters to analyze
the effects of viewpoint and motion diversity.
In all cases, we evaluate our models on real test videos.

\vspace{0.1cm}
\noindent\textbf{Real baselines.}
We start with our cross-view-subject protocol on NTU by training only with real data.
Table~\ref{table:ntubaselines} summarizes the results
of training the model on a single-view and testing on all views.
We observe a clear domain gap between different viewpoints, which can be naturally reduced
by adding more views in training. However, in the case when a single view is
available, this would not be possible.
If we train only with $0^\circ$,
the performance is high (83.9\%) when tested on
$0^\circ$, but significantly drops (42.9\%) when tested on $90^\circ$.
In the remaining of our experiments on NTU, we assume that
only the frontal viewpoint ($0^\circ$) is available.

\vspace{0.1cm}
\noindent\textbf{Non-uniform frame sampling.}
We note the consistent improvement of non-uniform frame sampling
over the uniform consecutive sampling
in all settings in Table~\ref{table:ntubaselines}.
Additional experiments about video frame sampling,
such as the optical flow stream, can be found in
\if\sepappendix1{the supplemental material.}
\else{Appendix~\ref{subsec:app:nonuniform}.}
\fi
We use our non-uniform sampling strategy for both
RGB and flow streams in the
remainder of experiments unless specified otherwise.

\vspace{0.1cm}
\noindent\textbf{Synth+Real training.}
Next, we report the improvements obtained by synthetically increasing view diversity.
We train the 60 action classes from NTU by combining
the real $0^\circ$ training data and the synthetic data
augmented from real with 8 viewpoints, i.e.~$0^\circ$:$45^\circ$:$360^\circ$.
Table~\ref{table:ntusynth} compares the results of Real, Synth, and Synth+Real trainings
for RGB and Flow streams,
as well as their combination.
The performance of the flow stream is generally
lower than that of the RGB stream,
possibly due to the fine-grained categories which
cannot be distinguished with coarse motion fields.

\begin{figure*}
    \centering
    \setlength{\tabcolsep}{0.15\linewidth}
    \begin{tabular}{ccc}
        {\footnotesize $0^\circ$} & {\footnotesize $45^\circ$} & {\footnotesize $90^\circ$}\\
    \end{tabular}
    \includegraphics[width=0.325\linewidth]{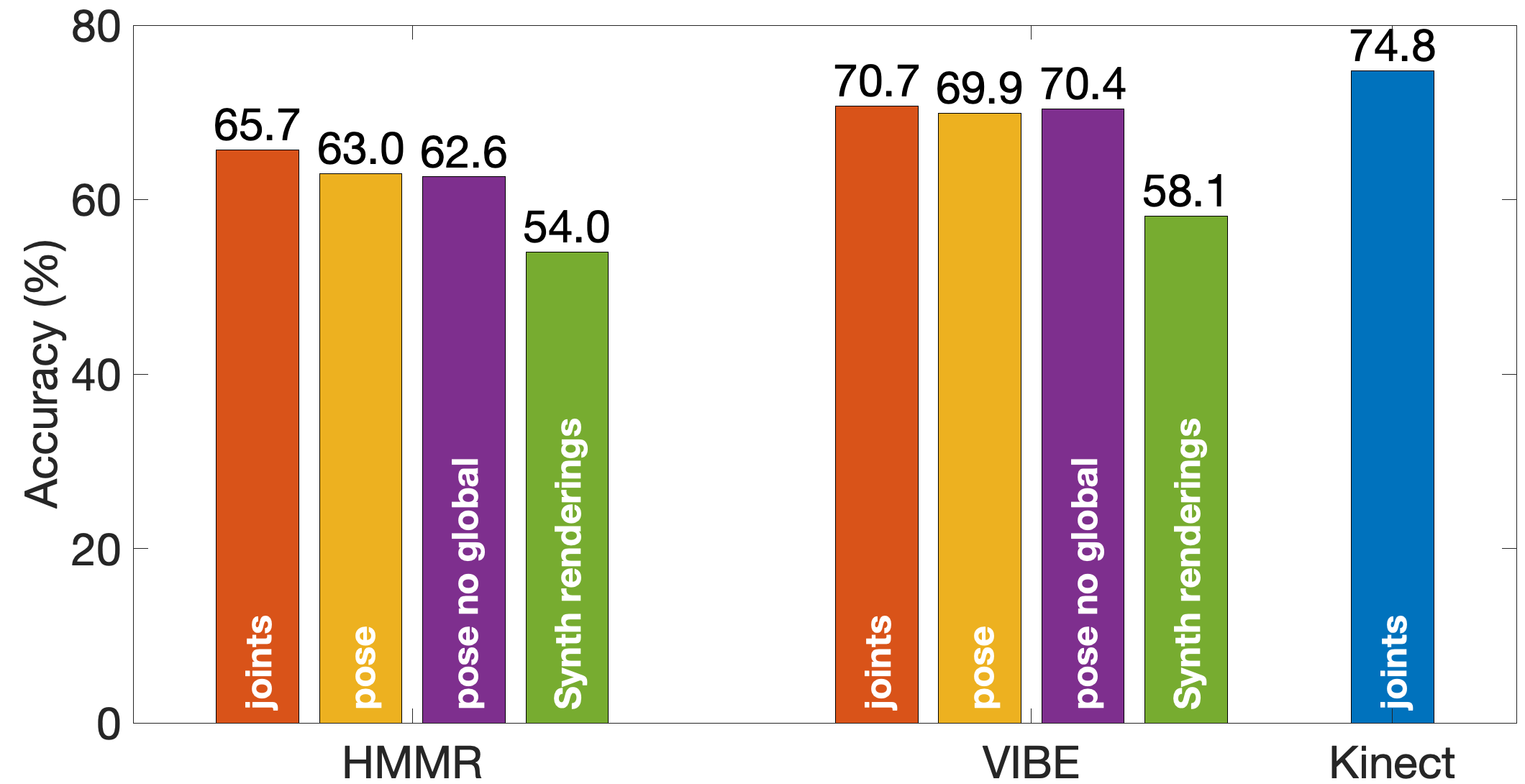}
    \includegraphics[width=0.325\linewidth]{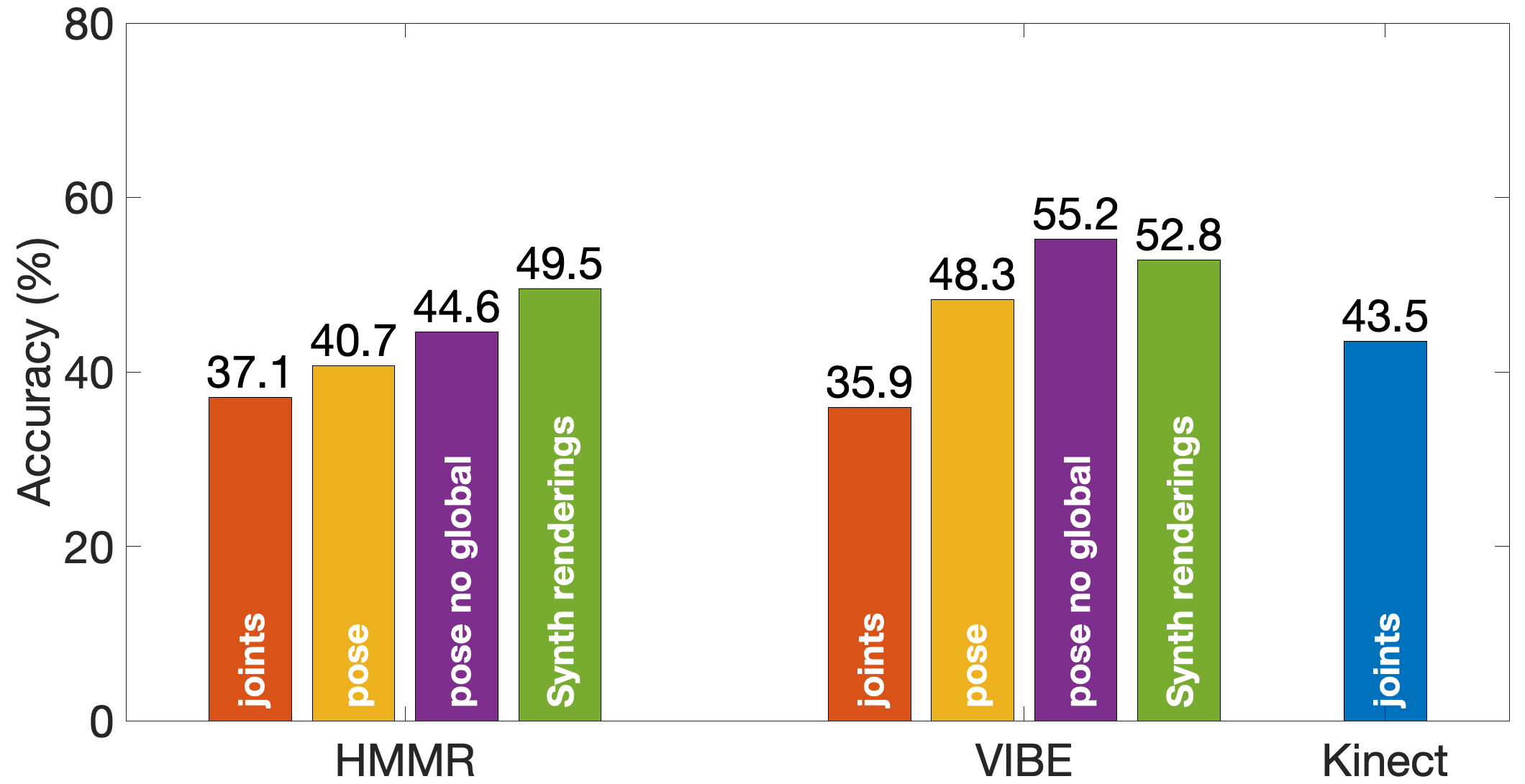}
    \includegraphics[width=0.325\linewidth]{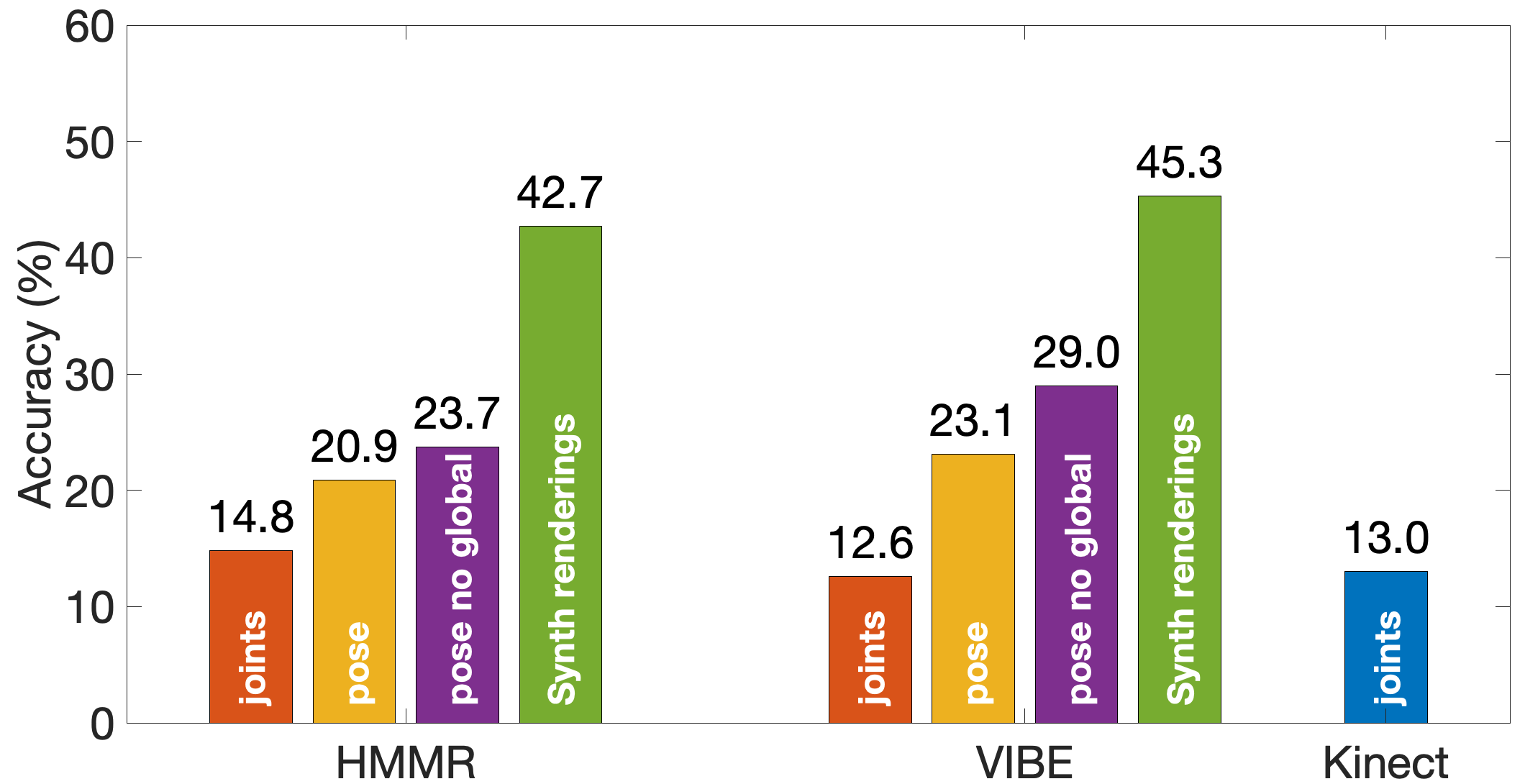}
    \caption{
        \cpfont 
        \textbf{Inputting raw motion parameters}
        performs significantly worse for the $90^\circ$ unseen viewpoint
        compared to synthetic renderings on the NTU CVS protocol. We compare various input representations
        with increasing view-independence (joint coordinates, SMPL pose parameters, SMPL pose parameters without the 
        global rotation). Experiments are carried out with SMPL model recovered with RGB-based methods 
        HMMR~\cite{humanMotionKZFM19} and VIBE~\cite{VIBECVPR2020},
        and depth-based Kinect joints. A 2D ResNet architecture 
        is used for motion parameter
        inputs similar to~\cite{KeCVPR17}. We also present an architecture
        study in Table~\ref{table:stgcn}.
        Note that significant gains are further possible when mixing the
        synthetic renderings with real videos. See text for interpretation.
    }
    \label{fig:input}
\end{figure*}

It is interesting to note that training only with synthetic data (Synth) reaches
63.0\% accuracy on real $0^\circ$ test data which indicates
a certain level of generalization capability from synthetic to real.
Combining real and synthetic training videos (Real+Synth),
the performance of the RGB stream increases from 53.6\% to 69.0\%
compared to only real training (Real),
on the challenging unseen $90^\circ$ viewpoint.
Note that the additional synthetic videos can be obtained
`for free', i.e.~without extra annotation cost.
We also confirm that even the noisy motion estimates
are sufficient to obtain significant improvements,
suggesting that the discriminative action information is still
present in our synthetic data.

The advantage of having a controllable data generation
procedure is to be able to analyze what components
of the synthetic data are important. In the following,
we examine a few of these aspects, such as quality
of the motion estimation, input representation, amount
of data, view diversity, and motion diversity.
Additional results can be found in
\if\sepappendix1{the supplemental material.}
\else{Appendix~\ref{sec:app:results}.}
\fi

\vspace{0.1cm}
\noindent\textbf{Quality of the motion estimation: HMMR vs VIBE.}
3D motion estimation from monocular videos has only recently
demonstrated convincing performance on unconstrained videos, opening up
the possibility to investigate our problem of action recognition
with synthetic videos. One natural question is whether the
progress in 3D motion estimation methods will improve the
synthetic data. To this end, we compare two sets of synthetic data,
keeping all the factors the same except the motion source:
Synth$_\mathit{HMMR}$
extracted with HMMR~\cite{humanMotionKZFM19}, Synth$_\mathit{VIBE}$ extracted
with VIBE~\cite{VIBECVPR2020}. Table~\ref{table:ntusynth} presents
the results. We observe consistent improvements with more accurate pose estimation
from VIBE over HMMR, suggesting that our proposed pipeline has great potential to
further improve with the 
progress in 3D recovery.

\vspace{0.1cm}
\noindent\textbf{Raw motion parameters as input.}
Another question is whether the motion estimation output, i.e.,
body pose parameters, can
be directly used as input to an action recognition model
instead of going through synthetic renderings.
We implement a simple 2D CNN architecture
similar to \cite{KeCVPR17} that inputs 16-frame pose sequence
in the form of 3D joint coordinates (24 joints for SMPL, 25 joints for Kinect)
or 3D joint rotations (24 axis-angle parent-relative rotations for SMPL, or 23
without the global rotation). In particular, we use a ResNet-18 architecture~\cite{He2016}.
We experiment with both HMMR and VIBE to use SMPL
parameters as input, as well as Kinect joints provided by the NTU dataset for comparison.
Figure~\ref{fig:input} reports the results of various pose representations
against the performance of synthetic renderings for three test views.
We make several observations: (i) Removing viewpoint-dependent factors,
e.g., pose parameters over joints, degrades performance on seen viewpoint, but consistently
improves on unseen viewpoints; (ii) Synthetic video renderings from all viewpoints significantly
improve over raw motion parameters for the challenging unseen viewpoint;
(iii) VIBE outperforms HMMR; (iv) Both RGB-based motion estimation methods
are competitive with the depth-based Kinect joints.

\begin{table}
    \centering
    \resizebox{0.89\linewidth}{!}{
        \begin{tabular}{llccc}
            \toprule
            Arch. & Input & $0^\circ$ & $45^\circ$ & $90^\circ$ \\
            \midrule
            2D ResNet & pose & 69.9 & 48.3 & 23.1 \\
            2D ResNet & pose no global & 70.4 & 55.2 & 29.0 \\
            \midrule
            ST-GCN & pose & 74.8 & 59.8 & 31.4 \\
            ST-GCN & pose no global & 75.6 & 60.9 & 36.2 \\
            \midrule
            3D ResNet & Synth & 58.1 & 52.8 & 45.3 \\
            \bottomrule
        \end{tabular}
    }
    \caption{
        \cpfont 
        \textbf{Architecture comparison:}
        We explore the influence of architectural improvements
        for pose-based action recognition models: 2D ResNet
        with temporal convolutions
        versus ST-GCN with graph convolutions on the SMPL pose parameters
        obtained by VIBE. While ST-GCN improves over 2D ResNet,
        the performance of the synthetic-only training with renderings
        remain superior for the unseen $90^\circ$ viewpoint.
    }
    \label{table:stgcn}
\end{table}

We note the significant boost with renderings (45.3\%) over pose parameters (29.0\%)
for the $90^\circ$ test view despite the same source of motion information for both.
There are three main differences which can be potential reasons.
First, the architectures 3D ResNet and 2D ResNet have different capacities.
Second, motion estimation from non-frontal viewpoints
can be challenging, negatively affecting the performance of pose-based methods, but
not affecting 3D ResNet (because pose estimation is not a required step).
Third, the renderings have the advantage that standard data augmentation techniques
on image pixels
can be applied, unlike the pose parameters which are not augmented.
More importantly, the renderings have the advantage
that they can be mixed with the real videos, which showed to
substantially improve the performance in Table~\ref{table:ntusynth}.

To explore the architectural capacity question,
we study the pose-based action recognition model further
and
experiment with the recent ST-GCN model \cite{stgcn} that makes use
of graph convolutions.
For this experiment, we use VIBE pose estimates
and compare ST-GCN with the 2D ResNet architecture in Table~\ref{table:stgcn}.
Although we observe improvements with using ST-GCN (29.0\% vs 36.2\%),
the synthetic renderings provide significantly better
generalization to the unseen 90$^\circ$ view (45.3\%).

\begin{figure*}
    \centering
    \setlength{\tabcolsep}{0.2\linewidth}
    \begin{tabular}{cc}
        {\footnotesize RGB} & {\footnotesize Flow} \\
    \end{tabular}
    \\
    \includegraphics[width=0.48\linewidth]{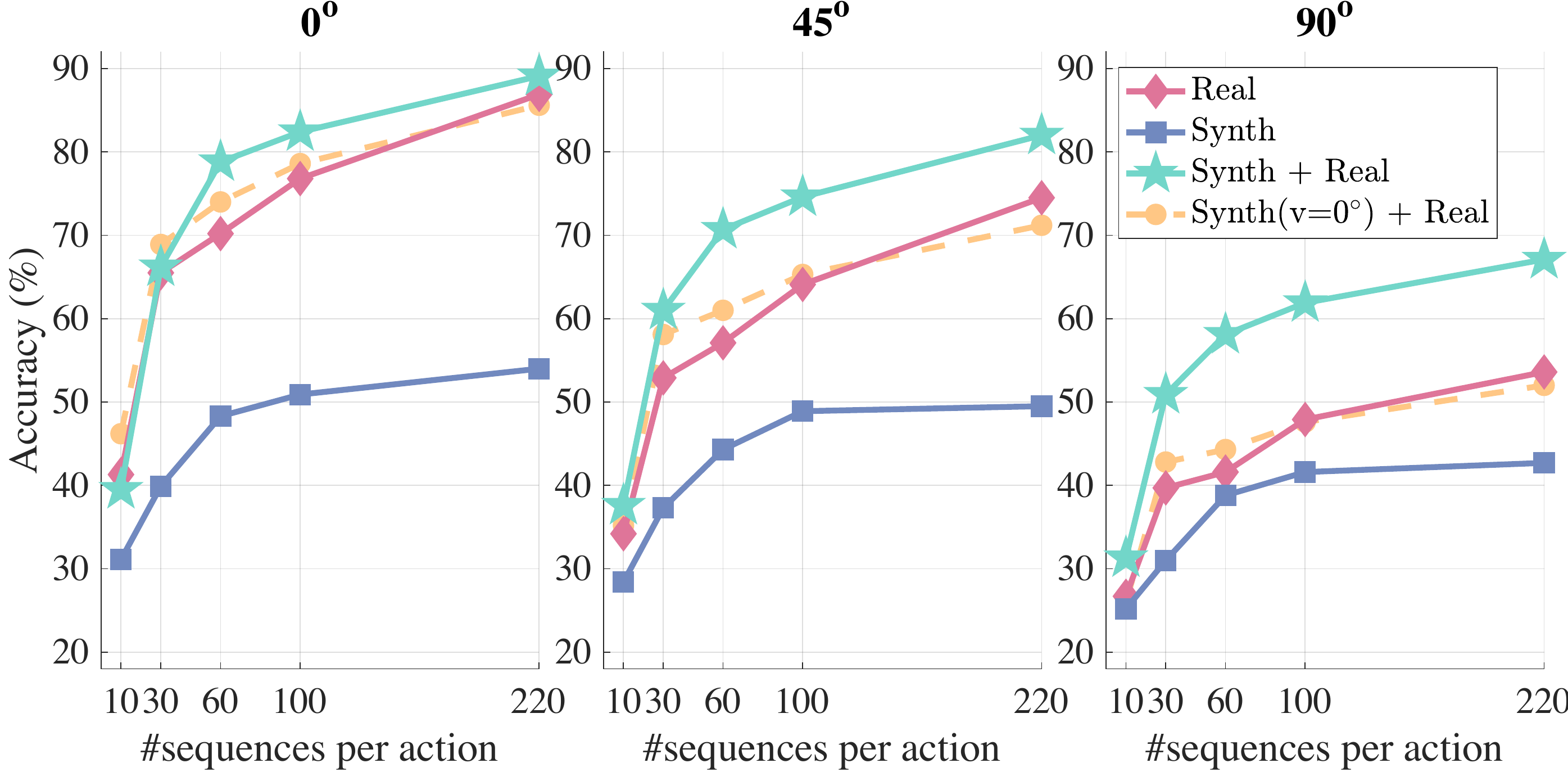} \hfill
    \includegraphics[width=0.48\linewidth]{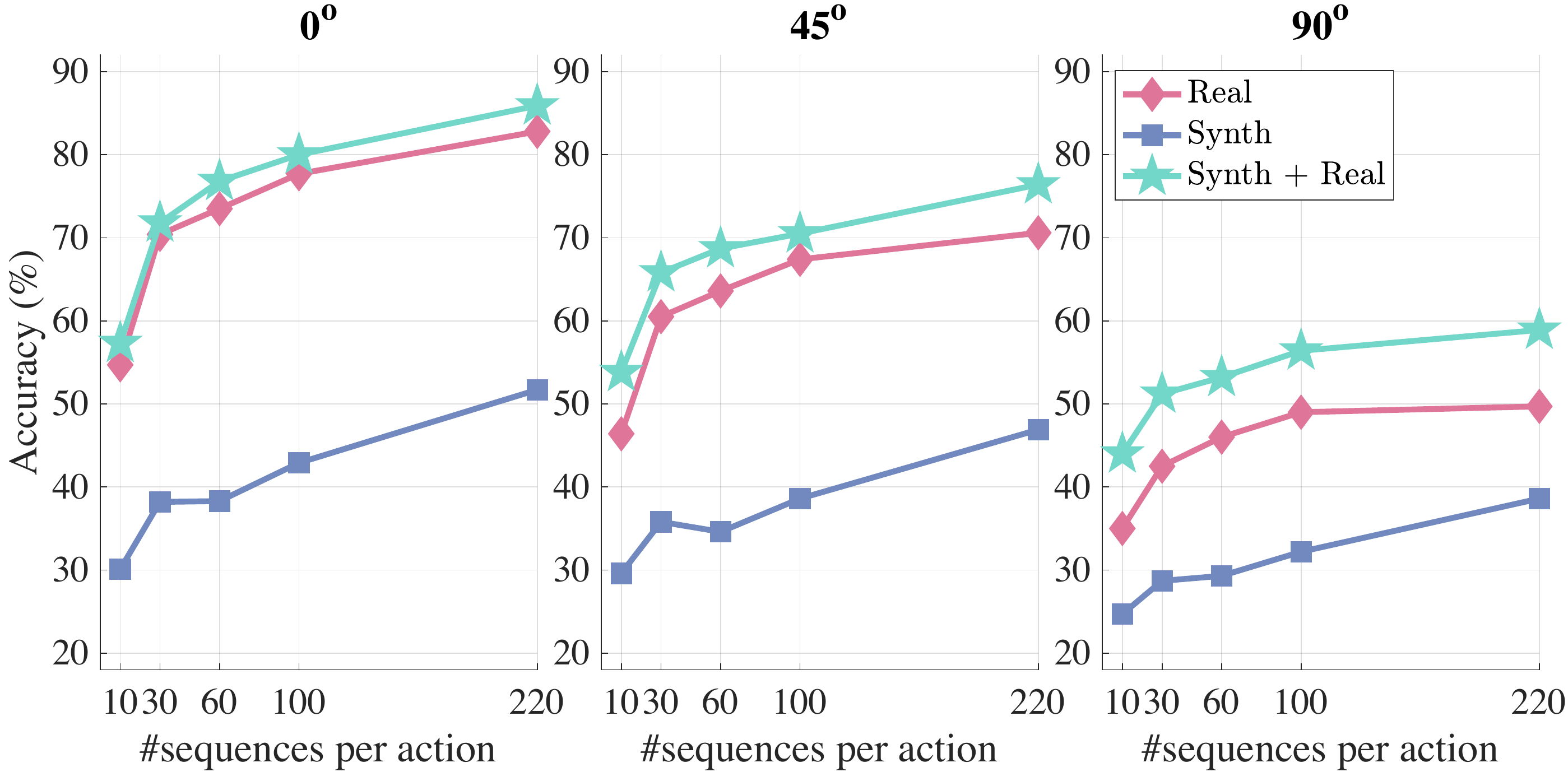}
    \caption{
        \cpfont 
        \textbf{Amount of data:}
        The number of real sequences per action %
        for: Real, Synth, Synth+Real training on NTU CVS split.
        Generalization to unseen viewpoints
        is significantly improved with the addition of synthetic data (green)
        compared to training only with real (pink).
        Real training contains the 0$^\circ$ view.
        We experiment with all 8 views (green) or only the 0$^\circ$ view (yellow)
        in the additional synthetic data.
        See text for interpretation.
    }
    \label{fig:synthamount}
\end{figure*}

\vspace{0.1cm}
\noindent\textbf{Amount of data.}
In the NTU CVS training split, we have about 220 sequences per action.
We take subsets with \{10, 30, 60, 100\} sequences per action,
and train the three scenarios:
Real, Synth, Synth+Real, for each subset.
Figure~\ref{fig:synthamount} plots the performance versus
the amount of data for these scenarios, for both RGB and Flow streams. We observe the
consistent improvement of complementary synthetic training,
especially for unseen viewpoints.
We also see that it is more effective to use synthetic data
at a given number of sequences per action.
For example, on the $90^\circ$ viewpoint,
increasing the number of sequences from 100 to 220 in the real data results
only in 4.6\% improvement (49.0\% vs 53.6\%, Real), while
one can synthetically augment the existing 100 sequences per action and obtain
64.7\% (Synth+Real)
accuracy without spending extra annotation effort. %

\vspace{0.1cm}
\noindent\textbf{View diversity.}
We wish to confirm that the improvements
presented so far are mainly due to the viewpoint variation
in synthetic data. The ``Synth(v=$0^\circ$)~+~Real'' plot
in Figure~\ref{fig:synthamount} indicates
that only the $0^\circ$ viewpoint from synthetic data is used.
In this case, we observe that the improvement is not consistent.
Therefore, it is important to augment viewpoints to obtain improvements.
Moreover, we experiment with having only $\pm45^\circ$ or $\pm90^\circ$
views in the synthetic-only training for 60 sequences per action.
In Table~\ref{table:synthviews},
we observe that the test performance is higher when
the synthetic training view matches the real test view.
However, having all 8 viewpoints at training benefits
all test views.

\begin{table}
    \centering
    \resizebox{0.89\linewidth}{!}{
        \begin{tabular}{lccc}
            \toprule
            & $0^\circ$                & $45^\circ$               & $90^\circ$               \\
            \midrule
            Synth($0^\circ$)           & \cellcolor{gray!25} 38.3 & 27.1                     & 17.9                     \\
            Synth($45^\circ$, $315^\circ$)      & 35.9                     & \cellcolor{gray!25} 34.2 & 26.8                     \\
            Synth($90^\circ$, $270^\circ$)      & 13.9                     & 18.3                     & \cellcolor{gray!25} 23.2 \\
            \midrule
            Synth($0^\circ$:$45^\circ$:$360^\circ$)       & \textbf{48.3}            & \textbf{44.3}            & \textbf{38.8}            \\
            \bottomrule
        \end{tabular}
    }
    \caption{
        \cpfont 
        \textbf{Viewpoint diversity:} The effect of the views in the synthetic training on the NTU CVS split.
        We train only with synthetic videos
        obtained from real data of 60 sequences per action.
        We take a subset of views from the synthetic data: 0$^\circ$, $\pm$45$^\circ$, $\pm$90$^\circ$.
        Even when synthetic, the performance
        is better when the viewpoints match between training and test (highlighted diagonal).
        The best performance is obtained with all 8 viewpoints combined.
    }
    \label{table:synthviews}
\end{table}

\begin{table}
    \centering
    \resizebox{0.99\linewidth}{!}{
        \begin{tabular}{cccccc}
            \toprule
            \#seq./ &           & motion        & \multicolumn{3}{c}{Test views}                \\ %
            action  & \#render & augm.   & $0^\circ$     & $45^\circ$    & $90^\circ$    \\ %
            \midrule
            10          & 1         & -              & 31.1          & 28.4          & 25.2          \\ %
            \midrule
            10          & 6         & -              & 33.1          & 31.6          & 26.2          \\ %
            10          & 6         & interp.  & 35.7          & 31.5          & 26.9          \\ %
            10          & 6         & \multicolumn{1}{l}{add. noise [frame]} & 25.0 & 24.2 & 21.4 \\ 
            10          & 6         & \multicolumn{1}{l}{add. noise [every 25f]} & 32.5 & 31.3 & 28.0  \\
            10          & 6         & \multicolumn{1}{l}{add. noise [video]} & \textbf{37.9} & \textbf{35.9} & \textbf{31.5} \\ %
            \midrule
            60          & 1         & -              & 48.3          & 44.3          & 38.8          \\ %
            \bottomrule
        \end{tabular}
    }
    \caption{
        \cpfont 
        \textbf{Motion diversity:} We study the effect of motion diversity in the synthetic training on a subset of the NTU CVS split. The results
        indicate that clothing, body shape diversity is not as important as motion
        diversity (second and last rows). We can significantly improve
        the performance by motion augmentations, especially
        with a video-level additive noise on the joint rotations (second and sixth rows).
        Here, each dataset is rendered with all 8 views and the training is only
        performed on synthetic data. At each rendering, we randomly sample clothing, body
        shape, lighting etc.}
    \label{table:synthmotions}
\end{table}

\begin{table*}%
    \begin{minipage}{.4\textwidth}
	\centering
	\resizebox{.99\linewidth}{!}{
        \begin{tabular}{cllc}
            \toprule
            \multicolumn{2}{l}{Method}                                        & Modality & Accuracy (\%) \\
            \midrule
            \multicolumn{2}{l}{VS-CNN~\cite{hri40}}                            & Skeleton & 29.0          \\
            \multicolumn{2}{l}{JOULE~\cite{joule2017} (by~\cite{hri40})}       & RGB      & 31.0          \\
            \multicolumn{2}{l}{ResNeXt-101~\cite{HaraCVPR2018} (by~\cite{hri40})} & RGB   & 32.0          \\
            \multicolumn{2}{l}{ResNeXt-101~\cite{HaraCVPR2018} (ours)} & RGB   & 45.2          \\
            \midrule
            \midrule
            RGB & Real [uniform]                              & RGB      & 36.1          \\
            \midrule
            \multirow{2}{*}{RGB} & Real & RGB      & 49.4          \\
            & Synth + Real                                     & RGB      & \textbf{66.4} \\
            \midrule
            \multirow{2}{*}{Flow} & Real & RGB      & 63.5             \\
            & Synth + Real                                     & RGB      & \textbf{73.1}    \\
            \midrule
            \multirow{2}{*}{RGB + Flow}& Real                                  & RGB      & 63.2             \\
            & Synth + Real                            & RGB      & \textbf{76.1}    \\
            \bottomrule
        \end{tabular}
	}
\end{minipage}
\begin{minipage}{.6\textwidth}
	\includegraphics[width=1\linewidth]{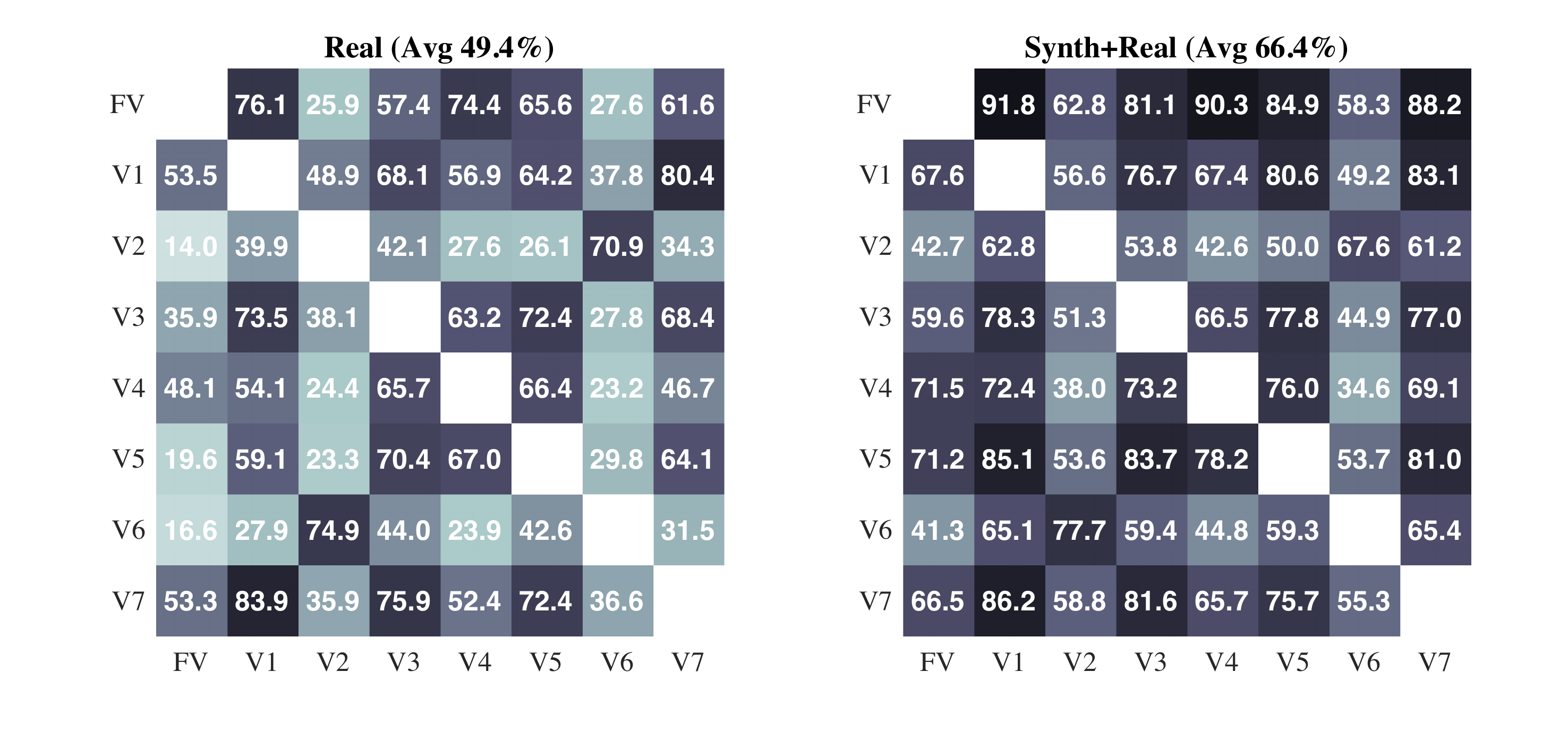}
    \end{minipage}
    \mbox{}\vspace{-.5cm}\\
	\caption{
        \cpfont 
        \textbf{UESTC dataset Cross View I protocol:} Training on 1 viewpoint and testing on all the others.
		The plots on the right show individual performances for the RGB networks. The rows and columns of the matrices correspond to training and testing views, respectively.
		We obtain significant improvements over the state of the art,
		due to our non-uniform frame sampling and synthetic training.}
    \mbox{}\vspace{-.9cm}\\
	\label{table:uestc_cv1}
\end{table*}

\begin{table*}
    \centering
    \resizebox{0.99\linewidth}{!}{
        \begin{tabular}{c lclllll|lllll|c}
            \toprule
            \multicolumn{2}{l}{Training views:} & & \multicolumn{5}{c|}{V1, V3, V5, V7} & \multicolumn{5}{c|}{FV, V2, V4, V6} & \\
            \midrule
            \multicolumn{2}{l}{Test views:} & & FV & V2 & V4 & V6 & Avg$_{even}$ & V1 & V3 & V5 & V7 & Avg$_{odd}$ & Avg \\
            \midrule
            \multicolumn{2}{l}{VS-CNN~\cite{hri40}} & Skeleton & 87.0 & 54.0 & 71.0 & 60.0 & 68.0 & 87.0 & 58.0 & 60.0 & 87.0 & 73.0 & 70.5\\
            \multicolumn{2}{l}{JOULE~\cite{joule2017} (by~\cite{hri40})} & RGB & 74.0 & 49.0 & 57.0 & 55.0 & 58.8 & 74.0 & 48.0 & 47.0 & 80.0 & 62.3 & 60.6 \\
            \multicolumn{2}{l}{ResNeXt-101~\cite{HaraCVPR2018} (by~\cite{hri40})} & RGB & 51.0 & 40.0 & 54.0 & 39.0 & 46.0 & 52.0 & 44.0 & 48.0 & 52.0 & 49.0 & 47.5\\
            \multicolumn{2}{l}{ResNeXt-101~\cite{HaraCVPR2018} (ours)} & RGB & 78.0 & 71.7 & 79.4 & 65.4 & 73.6 & 94.0 & 91.3 & 89.9 & 89.9 & 91.3 & 82.5 \\
            \midrule
            \midrule
            RGB & Real [uniform] & RGB & 78.1 & 63.1 & 76.6 & 46.4 & 66.1 & 92.1 & 80.9 & 85.5 & 86.1 & 86.2 & 76.1 \\
            \midrule
            \multirow{2}{*}{RGB} & Real & RGB & 69.9 & 57.1 & 79.1 & 51.1 & 64.3 & 91.3 & 86.8 & 89.4 & 88.9 & 89.1 & 76.7 \\
            & Synth + Real & RGB & \textbf{79.4} & \textbf{75.8} & \textbf{83.6} & \textbf{73.3} & \textbf{78.0} & \textbf{95.8} & \textbf{91.2} & \textbf{92.8} & \textbf{93.9} & \textbf{93.4} & \textbf{85.7} \\
            \midrule
            \multirow{2}{*}{Flow} & Real & RGB & 73.5 & 68.4 & 81.2 & 60.3 & 70.9 & 94.6 & 83.4 & \textbf{89.8} & 90.8 & 89.7 & 80.3 \\
            & Synth + Real & RGB & \textbf{79.0} & \textbf{73.1} & \textbf{84.6} & \textbf{73.7} & \textbf{77.6} & \textbf{95.2} & \textbf{87.8} & 89.2 & \textbf{92.7} & \textbf{91.2} & \textbf{84.4} \\
            \midrule
            \multirow{2}{*}{RGB + Flow} & Real & RGB & 74.5 & 68.4 & 82.4 & 59.4 & 71.2 & 95.8 & 88.9 & 91.4 & 92.3 & 92.1 & 81.7 \\
            & Synth + Real & RGB & \textbf{82.4} & \textbf{77.7} & \textbf{85.6} & \textbf{76.8} & \textbf{80.6} & \textbf{96.6} & \textbf{92.3} & \textbf{92.9} & \textbf{94.9} & \textbf{94.2} & \textbf{87.4} \\
            \bottomrule
        \end{tabular}
    }
    \caption{
        \cpfont 
        \textbf{UESTC dataset Cross View II protocol:} Training on 4 odd viewpoints,
        testing on 4 even viewpoints (left), and vice versa (right). We present the results
        on both splits and their average for the RGB and Flow streams, as well as the RGB+Flow late fusion.
        Real+Synth training consistently outperforms the Real baseline.}
    \label{table:uestc_cv2}
\end{table*}

\vspace{0.1cm}
\noindent\textbf{Motion diversity.}
Next, we investigate the question whether motions can be
diversified and whether this is beneficial for synthetic training.
There are very few attempts towards this direction~\cite{DeSouza:Procedural:CVPR2017}
since synthetic data has been mainly used for static images.
Recently, \cite{Liu2019Temp} introduced interpolation
between distinct poses to create new poses in synthetic data
for training %
3D pose estimation;
however, its contribution over existing poses was not experimentally validated.
In our case, we need to preserve the action information,
therefore, we cannot generate unconstrained motions.
Generating realistic motions is a challenging
research problem on its own and is out of the scope of this paper.
Here, we experiment with motion augmentation to increase
diversity.

As explained in Section~\ref{subsec:approach:estimation},
we generate new motion sequences by
(i) interpolating
between motion pairs of the same class,
or by (ii) additive noise on the pose parameters.
Table~\ref{table:synthmotions} presents the results
of this analysis when we train only with synthetic data
and test on the NTU CVS protocol. We compare to the baseline where 10 motion
sequences per action are rendered once per viewpoint (the first row).
We render the same sequences without motion augmentation
6 times (the second row)
and obtain marginal improvement.
On the other hand, having 60 real motion sequences per action
significantly improves (last row) and is our upper bound
for the motion augmentation experiments.
That means that the clothing, body shape, lighting,
i.e.~appearance diversity is not as important as motion diversity.
We see that generating new sequences with
interpolations improves over the baseline.
Moreover, perturbing the joint rotations across the video
with additive noise is simple and effective, with
performance increase of about 5\% (26.2\% vs 31.5\%) over rendering 6 times
without motion augmentation.
To justify the video-level noise (i.e., one value to add
to all frames), in Table~\ref{table:synthmotions}, we also 
experiment with frame-level noise
and a hybrid version where we independently sample a noise at every 25 
frames, which are interpolated for the frames in between. These
renderings qualitatively remain very noisy, reducing the performance
in return.

\begin{table}
    \centering
    \resizebox{0.99\linewidth}{!}{
        \begin{tabular}{r@{\hspace{0.1cm}}llcc}
            \toprule
            Method                            &                    & Modality                             & CS   & CV   \\
            \midrule
            Shahroudy~\cite{NTURGBD}          & Part-LSTM          & Skeleton                             & 62.9 & 70.3 \\ %
            Liu~\cite{LiuECCV2016}            & ST-LSTM            & Skeleton                             & 69.2 & 77.7 \\ %
            Liu~\cite{LiuCVPR2017}            & GCA-LSTM           & Skeleton                             & 74.4 & 82.8 \\ %
            Ke~\cite{KeCVPR17}                & MTLN               & Skeleton                             & 79.6 & 84.8 \\ %
            Liu~\cite{Liu2017ESV}             & View-invariant     & Skeleton                             & 80.0 & 87.2 \\ %
            Baradel~\cite{Baradel17}          & Hands attention    & RGB+Skeleton                         & 84.8 & 90.6 \\ %
            Liu~\cite{Liu_2018_CVPR}          & Pose evolution     & RGB+Depth                            & 91.7 & 95.3 \\ %
            Si~\cite{Si19}                    & Attention LSTM     & Skeleton                             & 89.2 & 95.0 \\ %
            Shi~\cite{Shi19twostream}         & 2s-AGCN            & Skeleton                             & 88.5 & 95.1 \\ %
            Shi~\cite{Shi19dg}                & DGNN               & Skeleton                             & 89.9 & 96.1 \\ %
            \midrule
            Baradel~\cite{Baradel17}          & Hands attention    & RGB \footnotesize{(Pose)}            & 75.6 & 80.5 \\ %
            Liu~\cite{Liu_2018_CVPR}          & Pose evolution     & RGB \footnotesize{(Pose)}            & 78.8 & 84.2 \\ %
            Zolfaghari~\cite{ZolfaghariOSB17} & Multi-stream       & RGB \footnotesize{(Pose+Flow)}       & 80.8 & -    \\ %
            Luvizon~\cite{Luvizon20182D3DPE}  & Multi-task         & RGB \footnotesize{(Pose)}            & 85.5 & -    \\ %
            Baradel~\cite{baradel18}          & Glimpse clouds     & RGB  \footnotesize{(Pose)}           & 86.6 & 93.2 \\ %
            Wang~\cite{Wang_2018_ECCV}        & DA-Net             & RGB \footnotesize{(Flow)}            & 88.1 & 92.0 \\ %
            Luo~\cite{luo_eccv18_graph}       & Graph distillation & RGB \footnotesize{(Pose+Flow+Depth)} & 89.5 & -    \\ %
            \midrule
            \multicolumn{2}{l}{\qquad \quad Real RGB [uniform]}         & RGB                                  & 86.3 & 90.8 \\
            \midrule
            \multicolumn{2}{l}{\qquad \quad Real RGB}                   & RGB                                  & 89.0 & 93.1 \\
            \multicolumn{2}{l}{\qquad \quad Real Flow}                  & RGB \footnotesize{(Flow)}            & 84.4 & 90.9 \\
            \multicolumn{2}{l}{\qquad \quad Real RGB+Flow}              & RGB \footnotesize{(Flow)}            & 90.0 & 94.3 \\
            \midrule
            \multicolumn{2}{l}{Synth+Real RGB}                          & RGB                                  & 89.6 & 94.1 \\
            \multicolumn{2}{l}{Synth+Real Flow}                         & RGB \footnotesize{(Flow)}            & 85.6 & 91.4 \\
            \multicolumn{2}{l}{Synth+Real RGB+Flow}                     & RGB \footnotesize{(Flow)}            & \textbf{90.7} & \textbf{95.0} \\
            \bottomrule
        \end{tabular}
    }
    \caption{
        \cpfont
        \textbf{State of the art comparison:} We report on the standard protocols of NTU for completeness.
        We improve previous RGB-based methods (bottom)
        due to non-uniform sampling and synthetic training. Additional cues extracted
        from RGB modality are denoted in parenthesis. We perform on par with
        skeleton-based methods (top) without using the Kinect sensor.
    }
    \label{table:ntusota}
\end{table}

\subsection{Comparison with the state of the art}
\label{subsec:experiments:sota}

In the following, we employ the standard protocols for
UESTC and NTU datasets, and compare our performance with other works.
Tables~\ref{table:uestc_cv1} and~\ref{table:uestc_cv2} compare
our results to the state-of-the-art methods reported by~\cite{hri40}
on the recently released UESTC dataset,
on CV-I and CV-II protocols.
To augment the UESTC dataset, we use the VIBE motion estimation method.
We outperform the RGB-based methods JOULE~\cite{joule2017} and
3D ResNeXt-101~\cite{HaraCVPR2018} by a large margin
even though we use a less deep 3D ResNet-50 architecture.
We note that we have trained the ResNeXt-101
architecture~\cite{HaraCVPR2018}
with our implementation and obtained better results than our ResNet-50
architecture (45.2\% vs 36.1\% on CV-I, 82.5\% vs 76.1\% on CV-II).
This contradicts the results reported in~\cite{hri40}.
We note that a first improvement can be attributed to our
non-uniform frame sampling strategy. Therefore, we report
our uniform real baseline as well. A significant performance boost
is later obtained by having a mixture of synthetic and real training data.
Using only RGB input, we obtain 17.0\% improvement on the challenging
CV-I protocol over real data (66.4 vs 49.4). Using both RGB and flow,
we obtain 44.1\% improvement over the state of the art (76.1 vs 32.0).
We also report on the even/odd test splits of the CV-II protocol
that have access to multi-view training data. The synthetic data
again shows benefits over the real baselines.
Compared to NTU, which contains object interactions that we do not simulate,
the UESTC dataset focuses more on the anatomic movements, such as body exercises.
We believe that these results convincingly demonstrate
the generalization capability of our efficient synthetic data
generation method to real body motion videos.

In Table~\ref{table:ntusota}, we compare our results to the
state-of-the-art methods on standard NTU splits.
The synthetic videos are generated using the HMMR motion
estimation method.
Our results on both splits achieve state-of-the-art performance only with the RGB modality.
In comparison, \cite{baradel18,Luvizon20182D3DPE,ZolfaghariOSB17} use pose information during training.
\cite{luo_eccv18_graph} uses
other modalities from Kinect such as depth and skeleton during training.
Similar to us, \cite{Wang_2018_ECCV} uses a two-stream approach.
Our non-uniform sampling boosts the performance. We have
moderate gains with the synthetic data for both RGB and flow streams,
as the real training set is already large and similar to the test set.

\subsection{One-shot training}
\label{subsec:experiments:oneshot}

\begin{table}
    \centering
    \resizebox{0.99\linewidth}{!}{
        \begin{tabular}{ll | ccc}
            \toprule
            & Synth                 & \multicolumn{3}{c}{Accuracy (\%)} \\
            Method & background            & RGB           & Flow              & RGB+Flow\\
            \midrule
            Chance & - & 6.7 & 6.7 & 6.7 \\
            Real (Nearest n.) & - & 8.6 & 13.1 & 13.9 \\
            \midrule
            Synth & Mini-Kinetics & 9.4 & 10.3 & 11.6 \\
            Real          & -                     & 26.2          & 20.6              & 28.4          \\
            Synth + Real  & LSUN           & 26.3          & 21.1              & 29.2          \\
            Synth + Real  & Mini-Kinetics  & \textbf{32.7} & \textbf{22.3}     & \textbf{34.6} \\
            \bottomrule
        \end{tabular}
    }
    \caption{
        \cpfont
        \textbf{One-shot Kinetics-15:}
        Real training data consists of 1 training sample per category, i.e., 15 videos.
        Random chance and nearest neighbor rows present baseline performances for this setup. 
        We augment each training video with 5 different viewpoints by synthetically rendering SMPL sequences extracted from real data (i.e., 75 videos), blended on random backgrounds
        from the Mini-Kinetics training videos and obtain 6.5\% improvement over training only with real data.
        For the last 4 rows, we train only the last linear layer of the ResNeXt-101 3D CNN model pre-trained on Mini-Kinetics 200
        classes.
    }
    \label{table:kinetics}
\end{table}

We test the limits of our approach on unconstrained
videos of the Kinetics-15 dataset.
These videos
are challenging for several reasons.
First, the 3D human motion estimation fails often
due to complex conditions such as motion blur,
low-resolution, occlusion, crowded scenes,
and fast motion. Second, there exist
cues about the action context that are difficult to simulate,
such as object interactions, bias towards certain clothing or
environments for certain actions.
Assuming that body motions alone, even when noisy,
provide discriminative information for actions,
we augment the 15 training videos of one-shot Kinetics-15
subset synthetically using HMMR (see Figure~\ref{fig:kinetics})
by rendering at 5 viewpoints (0$^\circ$, 30$^\circ$, 45$^\circ$, 315$^\circ$, 330$^\circ$).

We use a pre-trained feature extractor model
and only train a linear layer from the features
to the 15 classes. We observe over-fitting with
higher-capacity models due to limited one-shot training data.
We experiment with two pre-trained models,
obtained from \cite{crasto2019mars}: RGB and flow. The models follow the
3D ResNeXt-101 architecture from \cite{HaraCVPR2018}
and are pre-trained on Mini-Kinetics-200 categories with $16 \times 112 \times 112$
resolution with consecutive frame sampling.

In Table~\ref{table:kinetics} (top), we first
provide simple baselines: nearest neighbor with pre-trained
features is slightly above random chance (8.6\% vs 6.7\% for RGB).
Table~\ref{table:kinetics} (bottom) shows training linear layers.
Using only synthetic data
obtains poor performance (9.4\%).
Training only with real data on the other hand obtains 26.2\%,
which is our baseline performance.
We obtain $\sim$6\% improvement by adding synthetic data.
We also experiment with static background images from the LSUN dataset~\cite{yu_lsun}
and note the importance of realistic noisy backgrounds
for generalization to in-the-wild videos.

%% file: conclusions.tex
\section{Conclusions}
\label{sec:conclusions}

We presented an effective methodology
for automatically augmenting action recognition datasets
with synthetic videos. We explored the importance of
different variations in the synthetic data, such as
viewpoints and motions. Our %
analysis
emphasizes the question on how to diversify motions
within an action category.
We obtain significant improvements for action recognition
from unseen viewpoints and one-shot training.
However, our approach is limited by the
performance of the 3D pose estimation, which can fail
in cluttered scenes. Possible future directions include
action-conditioned generative models for motion sequences
and simulation of contextual cues for action recognition.

%% file: appendix.tex
\renewcommand{\thefigure}{A.\arabic{figure}} %
\setcounter{figure}{0} 
\renewcommand{\thetable}{A.\arabic{table}}
\setcounter{table}{0} 

\appendix

This appendix provides detailed explanations
for several components of
our approach (Section~\ref{sec:app:implementation}).
We also report complementary results
for %
synthetic training,
and our non-uniform frame sampling strategy
(Section~\ref{sec:app:results}).

\section{Additional details}
\label{sec:app:implementation}

\vspace{0.1cm}
\noindent\textbf{SURREACT rendering.}
We build on the implementation of \cite{varol2017surreal} and use the Blender software. We add support for multi-person images, for using estimated motion inputs, for systematic viewpoint rendering, and different sources for background images. We use the cloth textures released by \cite{varol2017surreal}, i.e., 361/90 female, 382/96 male textures for training/test splits, respectively. The resolution of the video frames is similarly 320x240 pixels.
For background images, we used 21567/8790 train/test images extracted from NTU videos, and 23034/23038 train/test images extracted from UESTC videos, by sampling a region outside of the person bounding boxes. The rendering code takes approximately 6 seconds per frame, for saving RGB, body-part segmentation and optical flow data. We parallelize the rendering over hundreds of CPUs to accelerate the data generation.

\begin{figure}[t]
    \centering
    \includegraphics[width=0.99\linewidth]{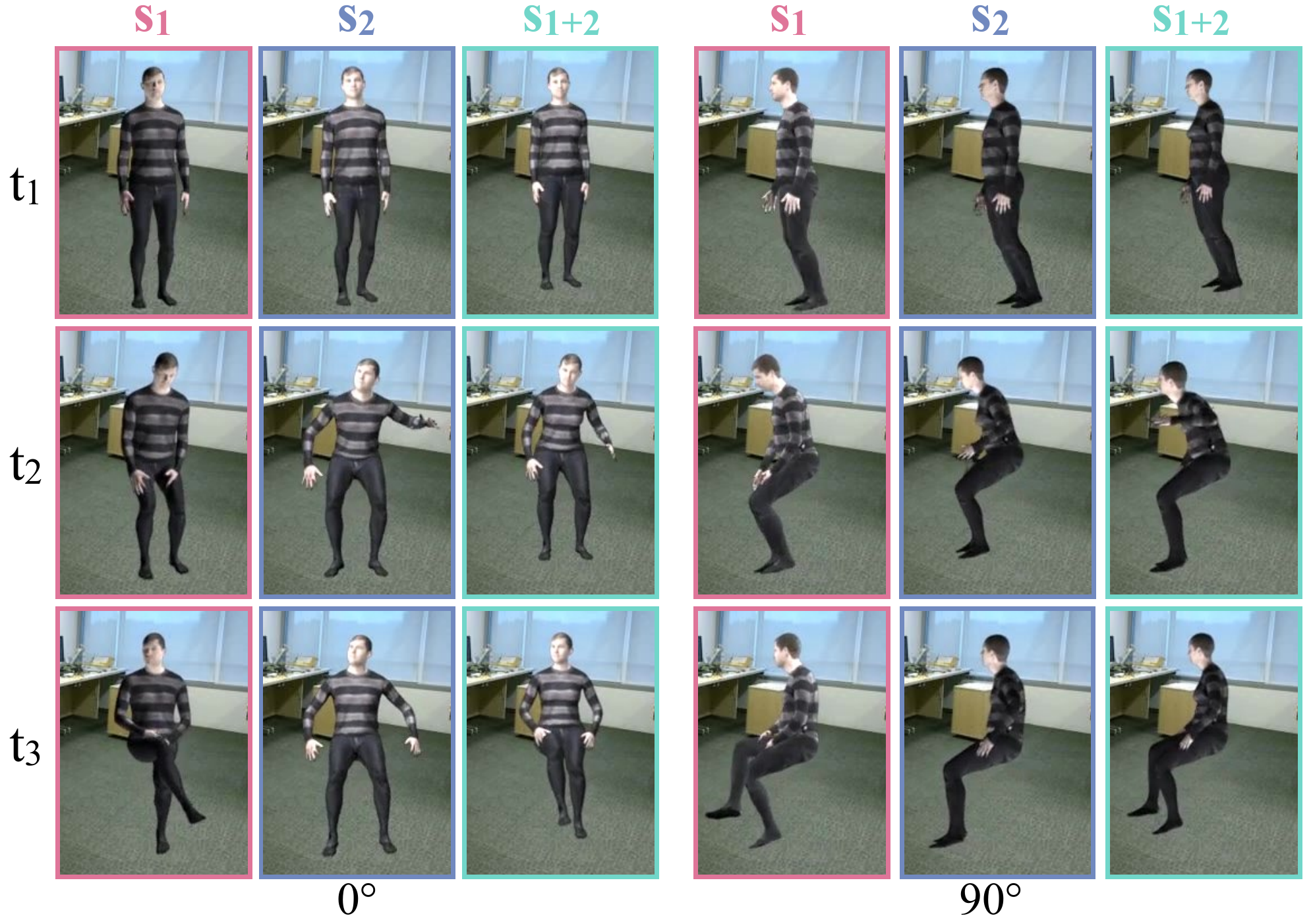}
    \caption{
        \cpfont
        Motion interpolation procedure for the \textit{sitting}
    action. Two temporally aligned sequences $s_1$ and $s_2$ from different individuals are interpolated
    to create $s_{1+2}$, from two viewpoints. Note the new arm and leg angles
    that contribute to motion diversity.}
    \label{fig:app:interpolation}
\end{figure}

\vspace{0.1cm}
\noindent\textbf{Motion sequence interpolation.}
As explained in
\if\sepappendix1{Section~3.2 of the main paper,}
\else{Section~\ref{subsec:approach:surreact} of the main paper,}
\fi
we explore creating new sequences
by interpolating pairs of motions from
the same action category. Here, we visually illustrate
this process. Figure~\ref{fig:app:interpolation}
shows two sequences of \textit{sitting down}
that are first aligned with dynamic time warping,
and then linearly interpolated. We only experiment
with equal weights when interpolating (i.e.~0.5), but one
can sample different weights when increasing
the number of sequences further.

\vspace{0.1cm}
\noindent\textbf{3D translation in SURREACT.}
\if\sepappendix1{In Section~3.2 of the main paper,}
\else{In Section~\ref{subsec:approach:surreact} of the main paper,}
\fi
we explained
that we translate the people in the $xy$ image plane
only when there are multiple people in the scene.
HMMR~\cite{humanMotionKZFM19} estimates the weak-perspective camera scale,
jointly with the body pose and shape.
We note that obtaining 3D translation of the person
in the camera coordinates is an ambiguous problem. It requires
the size of the person to be known. This becomes
more challenging in the case of multi-person videos.

HMMR relies on 2D pose estimation
to locate the bounding box of the person which then becomes the input
to a CNN. The CNN outputs a scale estimation $s_b$ together with the $[x_b, y_b]$ 
normalized image
coordinates of the person center with respect to the bounding box.
We first convert these values to be with respect to the original uncropped image:~$s$~and $[x, y]$.
We can recover an approximate value for the $z$ coordinate of the person center,
by assuming a fixed focal length $F=500$. The translation in $z$ then becomes:
$z = F / (0.5 * W * s)$, where $W$ is the image resolution %
and $s$ is the estimated camera scale.
The translation of the person center then becomes $[x, y, z]$. In practice,
the $z$ values are very noisy whereas $[x, y]$ values are more reliable.
We therefore assume that the person is always centered at $z=0$
and apply the translation only in the $xy$ plane.

\begin{table}
    \centering
    \resizebox{0.99\linewidth}{!}{
        \begin{tabular}{ll ccc }
            \toprule
            \#people & translation & $0^\circ$ & $45^\circ$ & $90^\circ$  \\
            \midrule
            single & xyz     & 18.3 & 17.8 & 15.0 \\
            multi  & xyz     & 21.0 & 21.2 & 17.3 \\
            multi  & xy      & 26.8 & 26.0 & 21.9 \\
            multi  & xy (when multi-person) & \textbf{28.5} & \textbf{27.2} & \textbf{23.0} \\
            \bottomrule
        \end{tabular}
    }
    \caption{
        \cpfont
        Training on different versions of the synthetic data
        generated from 10 sequences per action from the NTU CVS protocol.
        We train only on synthetic and test on the real test set.
        Multi-person videos in the synthetic training improves
        performance, especially in the interaction categories (see
        Figure~\ref{fig:app:confmatsingle}). The noisy translation estimates
        degrades generalization, therefore, we use only $xy$ translation
        and only in the case of multi-person. See text for further details.}
    \label{table:app:translation}
\end{table}

\begin{figure*}[t]
    \centering
    \includegraphics[width=0.99\linewidth]{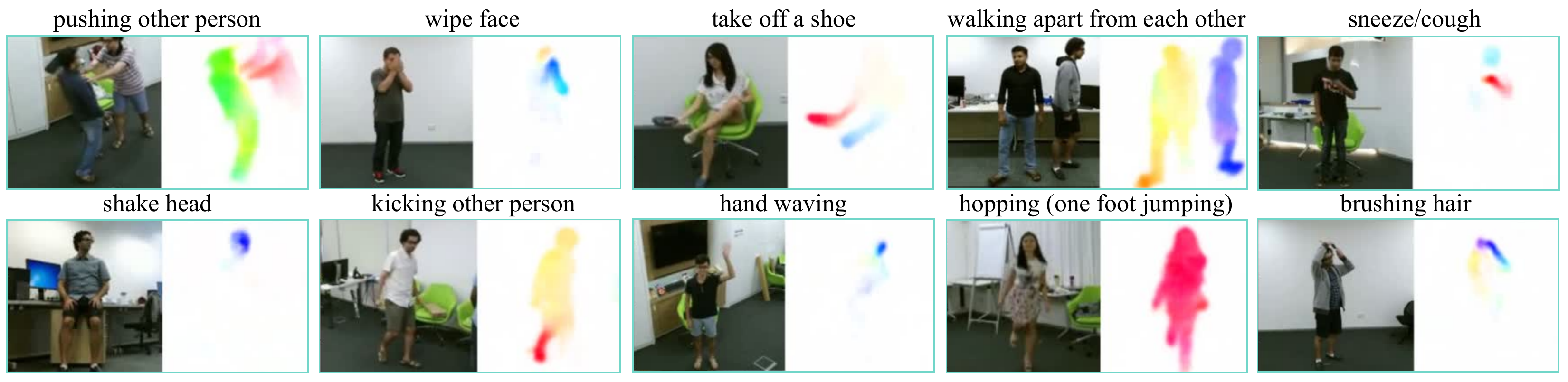}
    \caption{
        \cpfont
        Qualitative results for our optical flow estimation network trained on SURREACT, tested on the NTU dataset.}
    \label{fig:app:flowntu}
\end{figure*}

\begin{figure*}%
    \centering
    \includegraphics[width=0.99\linewidth]{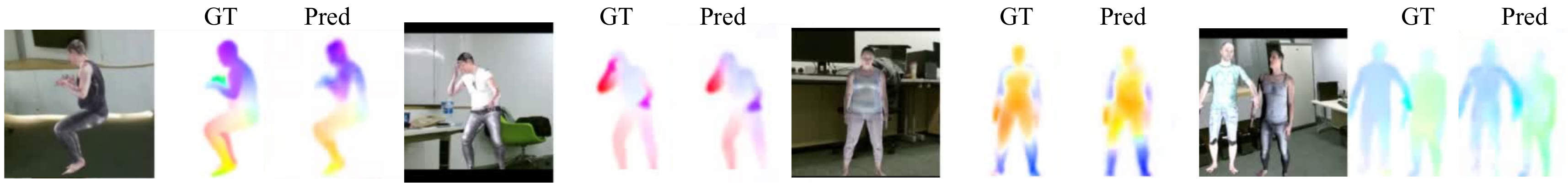}
    \caption{
        \cpfont
        Qualitative results for our optical flow estimation network trained and tested on SURREACT, together with the
        ground truth.}
    \label{fig:app:flowsynth}
\end{figure*}

We observe that due to the noisy 2D person detections
the estimated translation is noisy even in the $xy$ image plane,
leading to less generalization performance on real data
when we train only with synthetic data.
We validate this empirically in Table~\ref{table:app:translation}.
We render multiple versions of the synthetic dataset with 10 motion sequences
per action, each rendered from 8 viewpoints. We train only with this synthetic
data and evaluate on the real NTU CVS protocol. Including
multiple people improves performance (first and second rows),
mainly because 11 out of 60 action
categories in NTU are two-person interactions.
Figure~\ref{fig:app:confmatsingle} also shows
the confusion matrix of training only with single-person,
resulting in the confusion of the interaction categories.
Dropping the $z$ component from the translation further improves (second and third rows).
We also experiment with no translation if there is a single person,
and $xy$ translation only for the multi-person case (fourth row),
which has the best generalization performance.
This is unintuitive since some actions such as \textit{jumping}
are not realistic when the vertical translation is not simulated.
This indicates that the translation estimations from the real data
need further improvement
to be incorporated in the synthetic data. Our 3D CNN
is otherwise sensitive to the temporal jitter induced
by the noisy translations of the people.

\vspace{0.1cm}
\noindent\textbf{Flow estimation.}
We train our own optical flow estimation CNN,
which we use to compute the flow in an online fashion, during action
classification training. In other words, we do not require
pre-processing the videos for training.
To do so, we use a light-weight stacked hourglass
architecture~\cite{newell2016hourglass}
with two stacks. The input and output have $256 \times 256$
and $64 \times 64$ spatial resolution, respectively.
The input consists of 2 consecutive RGB frames of a video,
the output is the downsampled optical flow ground truth.
We train with mean squared error between the estimated
and ground truth flow values.
We obtain the ground truth from our synthetic SURREACT
dataset. Qualitative results of our optical flow estimates
can be seen in Figures~\ref{fig:app:flowntu} and ~\ref{fig:app:flowsynth}
on real and synthetic images, respectively.
When we compute the flow \textit{on-the-fly}
for action recognition, we loop over the 16-frame RGB
input to compute the flow between every 2 frames
and obtain 15-frame flow field as input to the action
classification network.

\vspace{0.1cm}
\noindent\textbf{Training details.}
We give additional details to
\if\sepappendix1{Section~3.3 of the main paper}
\else{Section~\ref{subsec:approach:training} of the main paper}
\fi
on the action classification training.
We train our networks for 50 epochs with an initial learning
rate of $10^{-3}$ which is decreased twice with a factor of $10^{-1}$
at epochs 40 and 45, respectively. For NTU, UESTC, and SURREACT datasets,
we spatially crop
video frames around the person bounding box
with random augmentations in scale and the center of the bounding box.
For the Kinetics dataset, we crop randomly with a bias towards the center.
We scale the RGB values between $[0, 1]$ and jitter the color channels
with a multiplicative coefficient randomly generated between $[0.8, 1.2]$ for each
channel. We subtract $0.5$ and clip the values between $[-0.5, 0.5]$
before inputting to the CNN.

\vspace{0.1cm}
\noindent\textbf{NTU CVS protocol.}
We provide Table~\ref{table:app:csvprotocol}
with the number of videos used for each NTU protocol,
summarizing the difference of our new CVS protocol
from the official cross-view (CV) and cross-subject (CS) splits.
The CVS protocol addresses two problems with the official splits: (i) while CV uses same subjects across splits and CS uses same views across splits, CVS uses different subjects and different views between train and test; (ii) our cross-view setup of train(0), test(90) is much more challenging than train(0+90), test(45) due to viewpoints being more distinct, especially a problem with CV where the same subjects are used in train and test. 

\begin{table}
    \centering
    \setlength{\tabcolsep}{4pt}
    \resizebox{0.69\linewidth}{!}{
        \begin{tabular}{l | ccc | l}
            \toprule
            & $0^\circ$ & $45^\circ$ & $90^\circ$ & Total \\
            \midrule
            Train subjects & 13386 & 13415 & 13288 & 40089 \\
            Test subjects & 5503 & 5517 & 5467 & 16487 \\
            \midrule
            Total & 18889 & 18932 & 18755 & 56576 \\
            \bottomrule
        \end{tabular}
    }
    \newline
    \vspace*{.2 cm}
    \newline
    \resizebox{0.99\linewidth}{!}{
        \begin{tabular}{ll | ccc | l | l}
            \toprule
             & & $0^\circ$ & $45^\circ$ & $90^\circ$ & Total \\
            \midrule
            \multirow{2}{*}{CS} & Train & 13225 & 13225 & 13225 & 39675 & Diff. sub., same view \\
            & Test & 5503 & 5517 & 5467 & 16487* & \\
            \midrule
            \multirow{2}{*}{CV} & Train & 18672 & - & 18672 & 37344 & Same sub., diff. view \\
            & Test & - & 18932 & - & 18932 & (easy: 0+90 train, 45 test) \\
            \midrule
            \multirow{2}{*}{CVS} & Train & 13225 & - & - & 13225 & Diff. sub., diff. view \\
            & Test & 5447 & 5447 & 5447 & 16341* & (challenging: 0 train, 90 test) \\
            \bottomrule
        \end{tabular}
    }
    \caption{
        \cpfont
        \textbf{Statistics of NTU splits:}
        The above table summarizes the partition of the dataset into views and subjects.
         The below table provides the number of videos we use for each NTU protocol.
        *The difference between these numbers is because we filter out some videos for which there exist no synchronized camera, to keep the number of test videos same for each test view. Similarly we use synchronized cameras in training, therefore slightly lower number of training videos in CV (37344 instead of 37644=18889+18755) and CS (39675 instead of 40089) but the test sets reflect the official list of videos.
    }
    \label{table:app:csvprotocol}
\end{table}

\section{Additional analyses}
\label{sec:app:results}
We analyze further the synthetic-only training (Section~\ref{subsec:app:synth})
and synthetic+real training (Section~\ref{subsec:app:synthreal}).
We define a synthetic test set and report the results of the models
in the main paper also on this test set. We present additional ablations.
We report the confusion matrix on the synthetic test
set, as well as on the real test set, which allows
us to gain insights about which action categories can
be represented better synthetically.
Finally, we explore the proposed non-uniform sampling
more in Section~\ref{subsec:app:nonuniform}.

\subsection{Synthetic-only training}
\label{subsec:app:synth}
Here, we define a synthetic test set based on the NTU actions,
and perform additional ablations on our synthetic data
such as different input modalities beyond RGB and flow,
effect of backgrounds, effect of further camera augmentations,
and confusion matrix analysis.

\begin{table}[t]
    \centering
    \resizebox{0.99\linewidth}{!}{
        \begin{tabular}{l|cccccccc|c}
            \toprule
            &\multicolumn{8}{c|}{Synth Test Views} \\
            & $0^\circ$                & $45^\circ$                & $90^\circ$               & $135^\circ$ & $180^\circ$ & $225^\circ$ & $275^\circ$              & $315^\circ$ &  Avg             \\
            \midrule
            0           & \cellcolor{gray!25} \textbf{81.7} & 41.7             & 10.0                     & 43.3        & 40.0        & 30.0        & 10.0                     & 50.0       & 38.3               \\
            45-315      & 58.3                     & \cellcolor{gray!25} 65.0  & 36.7                     & 53.3        & 43.3        & 48.3        & 50.0                     & \cellcolor{gray!25} 66.7 & 52.7 \\
            90-270      & 15.0                     & 35.0                      & \cellcolor{gray!25} 61.7 & 23.3        & 13.3        & 35.0        & \cellcolor{gray!25} 56.7 & 33.3             & 34.2        \\
            \midrule
            All 8       & 78.3            & \textbf{73.3}             & \textbf{61.7}            & \textbf{68.3} & \textbf{73.3} & \textbf{75.0} & \textbf{65.0}      & \textbf{71.7}    & \textbf{70.8}        \\
            \bottomrule
        \end{tabular}
    }
    \caption{
        \cpfont
        The performance of the view-augmented models
        from
        \if\sepappendix1{Table~3 of the main paper}
        \else{Table~\ref{table:synthviews} of the main paper}
        \fi
        on the synthetic test set.
        We train only with synthetic videos
        obtained from 60 sequences per action.
        We confirm that the viewpoints should match also for the
        synthetic test set. We report the viewpoint breakdown,
        as well as the average.}
    \label{table:app:synthviews}
\end{table}

\begin{table}[t]
    \centering
    \resizebox{0.89\linewidth}{!}{
        \begin{tabular}{cccc}
            \toprule
            \#sequences &           & motion         & Synth        \\
            per action  & \#renders & augmentation   &  All          \\
            \midrule
            10          & 1         & -              & 55.4          \\
            \midrule
            10          & 6         & -              & 55.0          \\
            10          & 6         & interpolation  & \textbf{58.8} \\
            10          & 6         & additive noise & 57.7          \\
            \midrule
            60          & 1         & -              & 70.8          \\
            \bottomrule
        \end{tabular}
    }
    \caption{
        \cpfont
        The performance of the motion-augmented models
    from
    \if\sepappendix1{Table~4 of the main paper}
    \else{Table~\ref{table:synthmotions} of the main paper}
    \fi
    on the synthetic test set.
    We train only with synthetic videos
    obtained from 60 sequences per action.
    Both augmentation approaches improve over the baseline.}
    \label{table:app:synthmotions}
\end{table}

\vspace{0.1cm}
\noindent\textbf{Synthetic test set.}
Similar to SURREAL~\cite{varol2017surreal},
we separate the assets such as cloth textures, body shapes,
backgrounds into train and test splits, which allows us
to validate our experiments also on a synthetic test set.
Here, we use one sequence per action from the real 0$^\circ$
test set to generate a small synthetic test set, i.e.~60 motion
sequences in total, rendered for the 8 viewpoints,
using the test set assets.

We report the performance of our models from
\if\sepappendix1{Tables~3~and~4 of the main paper}
\else{Tables~\ref{table:synthviews}~and~\ref{table:synthmotions} of the main paper}
\fi
on this set.
Table~\ref{table:app:synthviews} confirms that the viewpoints
should match between training and test for best results. Augmenting
with all 8 views benefits the overall results.
Table~\ref{table:app:synthmotions} presents the gains
obtained by motion augmentations on the synthetic test set.
Both interpolations and the additive noise improves
over applying no augmentation.

\vspace{0.1cm}
\noindent\textbf{Different input types.}
The advantage of having a synthetic dataset is to be able to
perform experiments with different modalities.
Specifically, we have ground-truth optical flow, body-part segmentation
for each video. We compare training with these input modalities
as opposed to RGB, or the estimated flow in Table~\ref{table:app:synthinputs}.
We evaluate on the real NTU CSV test set when applicable,
and on the synthetic test set. We see that
even when ground truth, the optical flow performs
worse than RGB, indicating difficulty of distinguishing
fine-grained actions only with flow fields.
Body-part segmentation on the other hand, outperforms
other modalities due to providing precise locations
for each body part and an abstraction which
reduces the gap between the training and test splits.
In other words, body-part segmentation is independent
of clothing, lighting, background effects, but only
contains motion and body shape information.
This result highlights that we can improve
action recognition by improving body part segmentation
as in~\cite{ZolfaghariOSB17}.

\begin{table}[t]
    \centering
    \resizebox{0.89\linewidth}{!}{
        \begin{tabular}{lccc | c }
            \toprule
            & \multicolumn{3}{c|}{Real}                     & Synth         \\
            Input type                & $0^\circ$     & $45^\circ$    & $90^\circ$    & All           \\
            \midrule
            Flow (Pred)               & 38.3          & 34.6          & 29.3          & 58.4          \\
            Flow (GT)                 & -             & -             & -             & 61.2          \\
            RGB                       & 48.3          & 44.3          & 38.8          & 70.8          \\
            Body-part segm (GT)       & -             & -             & -             & \textbf{71.7} \\
            \bottomrule
        \end{tabular}
    }
    \caption{
        \cpfont
        Different input types when training only with synthetic data and testing on 
        the synthetic test set, as well as the real NTU CVS test set when applicable. The
        data is generated from 60 sequences per action. The results indicate
        that body part segmentation can be an informative representation for action classification.
        Optical flow, even when ground truth (GT) is used, is less informative for fine-grained
        action classes in NTU.}
    \label{table:app:synthinputs}
\end{table}

\begin{table}[t]
    \centering
    \resizebox{0.89\linewidth}{!}{
        \begin{tabular}{rl ccc | c }
            \toprule
            &                        & \multicolumn{3}{c|}{Real}                     & Synth         \\
            \multicolumn{2}{c}{Backgrounds} & $0^\circ$     & $45^\circ$    & $90^\circ$    & All           \\
            \midrule
            random & LSUN                   & 39.1            & 37.3          & 32.5          & 70.8          \\
            random & NTU                    & 42.7          & 39.8          & 34.3          & 67.9          \\
            fixed  & NTU                    & \textbf{48.3} & \textbf{44.3} & \textbf{38.8} & \textbf{70.8} \\
            \bottomrule
        \end{tabular}
    }
    \caption{
        \cpfont
        Effect of synthetic data backgrounds
        for synthetic-only training. Results are reported
        both on the real NTU CVS set and the synthetic test set.
        The synthetic training is generated from 60 sequences per action.
        Matching the target background statistics improves generalization to real.
        See text for details.}
    \label{table:app:backgrounds}
\end{table}

\vspace{0.1cm}
\noindent\textbf{Effect of backgrounds.}
As explained in
\if\sepappendix1{Section~3.2 of the main paper,}
\else{Section~\ref{subsec:approach:surreact} of the main paper,}
\fi
we use 2D background images
from the target action recognition domain in our
synthetic dataset. We perform an experiment whether
this helps on the NTU CVS setup. The NTU dataset
is recorded in a lab environment, therefore
has specific background statistics. We train
models by replacing the background pixels of our
synthetic videos randomly by LSUN~\cite{varol2017surreal,yu_lsun}
images or the original NTU images outside the
person bounding boxes.
Table~\ref{table:app:backgrounds} summarizes the results.
Using random NTU backgrounds outperform using
random LSUN backgrounds. However, we note that
the process of using the segmentation mask
creates some unrealistic artifacts around the person,
which might contribute to the performance degradation.
We therefore use the fixed backgrounds from the
original renderings in the rest of the experiments.

\begin{figure}[t]
    \centering
    \includegraphics[width=0.99\linewidth]{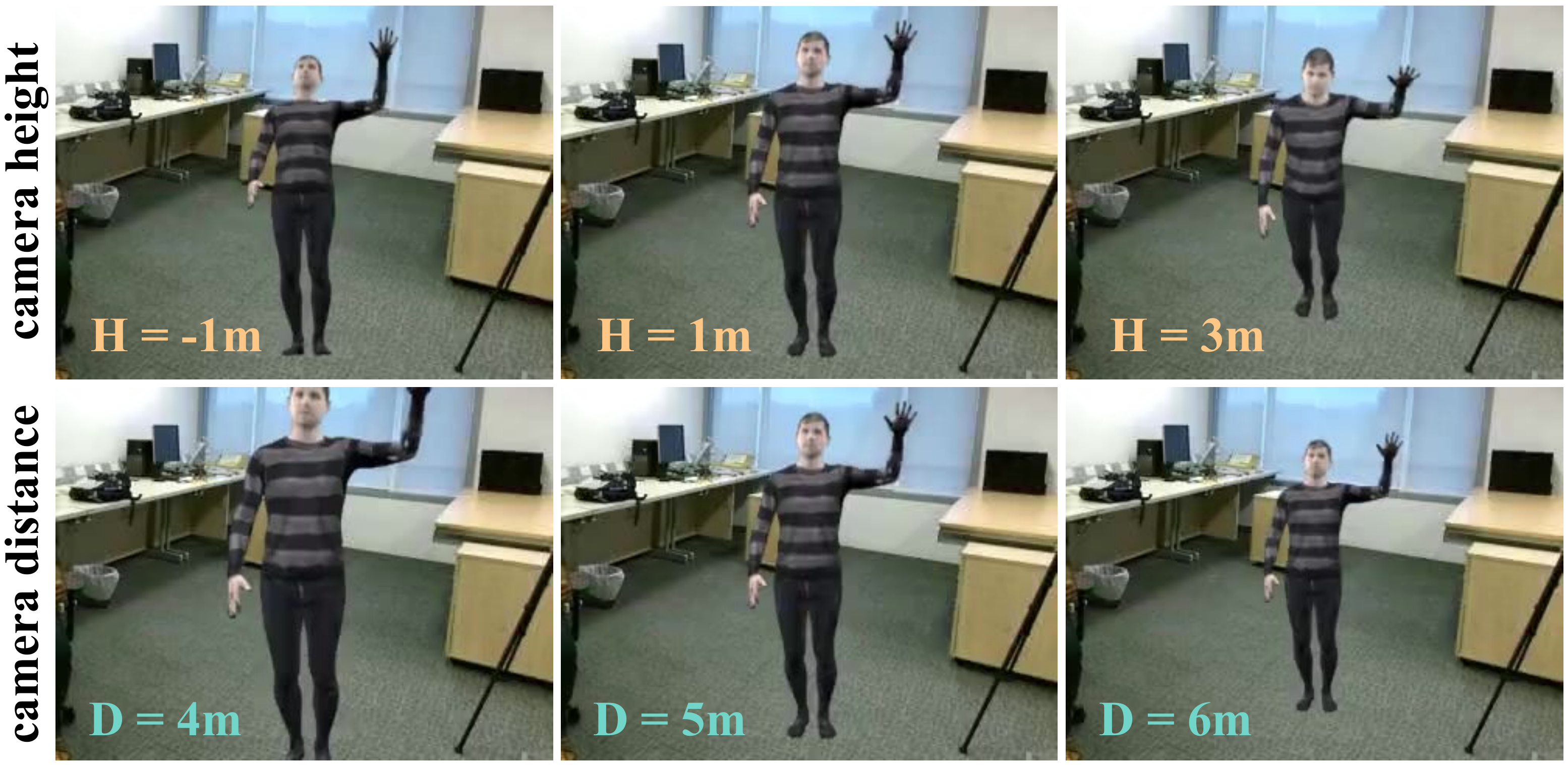}
    \caption{
        \cpfont
        We illustrate the limits for the camera height and distance parameters
        in SURREACT. We randomly sample between [-1, 3] and [4, 6] meters
        for the height and distance, respectively.}
    \label{fig:app:camera}
\end{figure}

\vspace{0.1cm}
\noindent\textbf{Effect of camera height/distance augmentations.}
As stated in
\if\sepappendix1{Section~3.2 of the main paper,}
\else{Section~\ref{subsec:approach:surreact} of the main paper,}
\fi
we randomize the height and the distance of the camera
to increase the viewpoint diversity within a certain azimuth rotation.
We evaluate the importance of this with a controlled
experiment in Table~\ref{table:app:camera}.
We render two versions of the synthetic training set
with 10 sequences per action from 8 viewpoints.
The first one has a fixed distance and height at 5 meters and 1 meter, respectively.
In the second one, we randomly sample from $[4, 6]$ meters for the distance,
and $[-1, 3]$ meters for the height. We see that
the generalization to real NTU CVS dataset is improved with increased randomness
in the synthetic training. Visuals corresponding to the pre-defined
range can be found in Figure~\ref{fig:app:camera}.

\begin{table}[t]
    \centering
    \resizebox{0.89\linewidth}{!}{
        \begin{tabular}{c ccc }
            \toprule
            camera height \& distance & $0^\circ$     & $45^\circ$    & $90^\circ$    \\
            \midrule
            fixed                     & 28.5          & 27.2          & 23.0          \\
            random                    & \textbf{31.3} & \textbf{28.1} & \textbf{24.3} \\
            \bottomrule
        \end{tabular}
    }
    \caption{
        \cpfont
        Training on different versions of the synthetic data
        generated from 10 sequences per action from the NTU CVS protocol.
        We ablate the importance of augmentations of the camera height
        and distance.
        We train only on synthetic and test on the real test set.
        We observe improvements with randomized camera positions
        besides the azimuth rotation.}
    \label{table:app:camera}
\end{table}

\begin{figure*}
    \centering
    \begin{tabular}{cc}
    Train: \textbf{Synth}, Test: \textbf{Synth} & 
    Train: \textbf{Synth}, Test: \textbf{Real (0$^\circ$)} \\
    \includegraphics[width=0.47\linewidth]{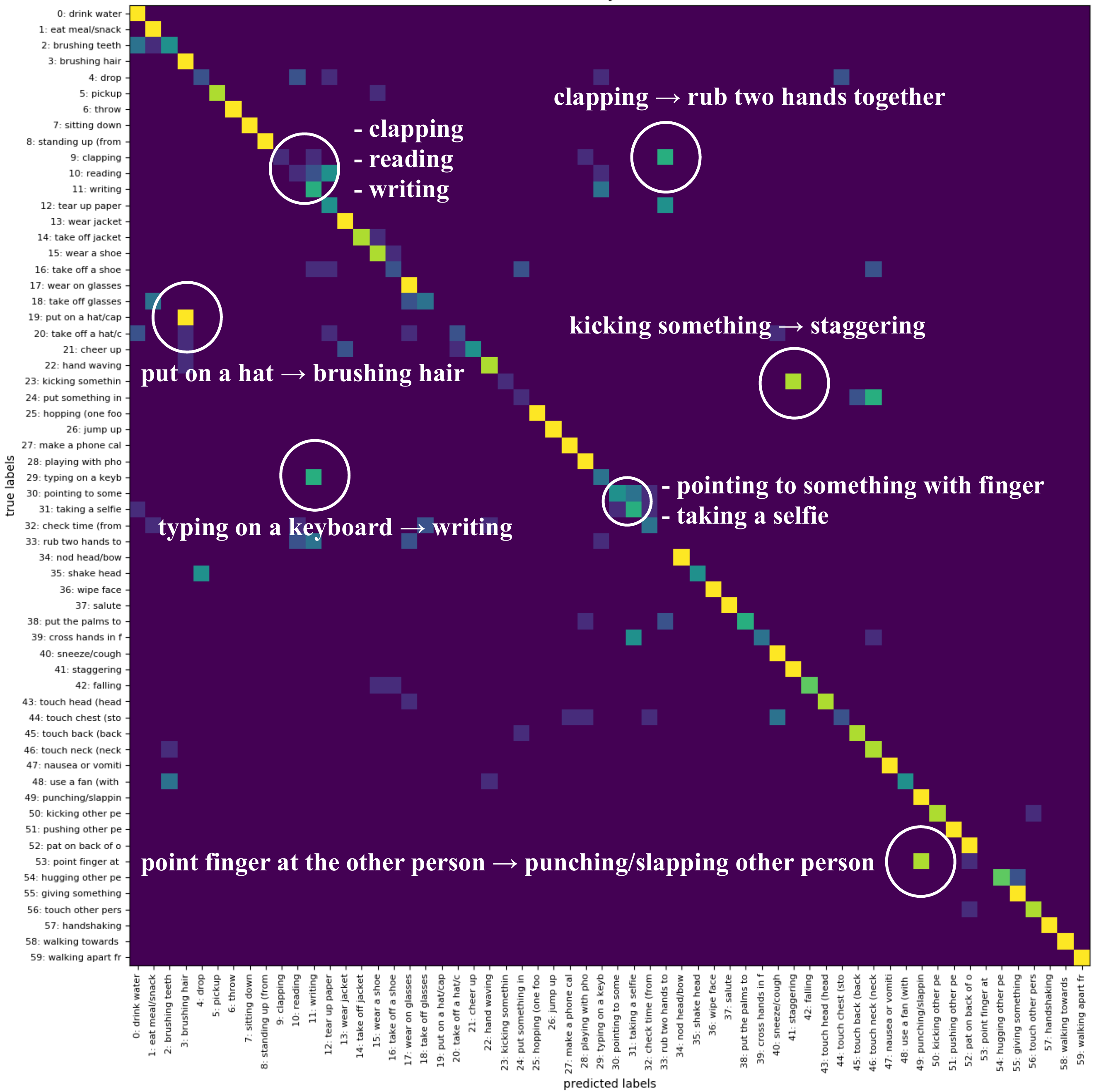} & 
    \includegraphics[width=0.47\linewidth]{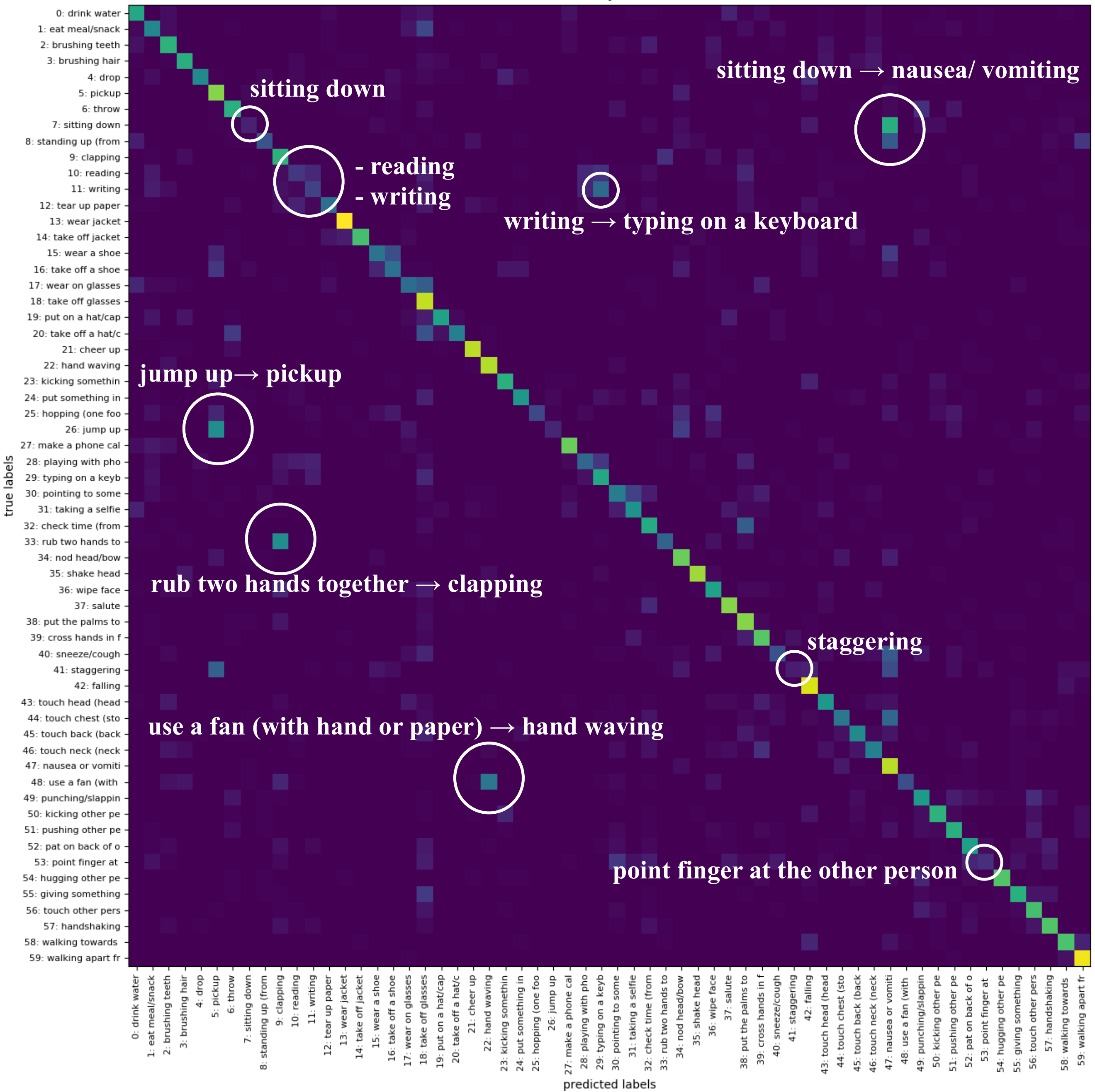} \\
    \end{tabular}
    \caption{
        \cpfont
        The confusion matrices for training only with the final synthetic
        data with all the 220 sequences per action. Both on the synthetic
        test set (left) and the real 0$^\circ$ view test (right),
        the confusions are often between classes that are characterized
        by fine-grained body movements and object-interaction classes.
    }
    \label{fig:app:confmat}
\end{figure*}

\begin{figure}
    \centering
    {\scriptsize Train: \textbf{Synth (single-person)}, Test: \textbf{Real(0$^\circ$)}} \\
    \includegraphics[width=0.99\linewidth]{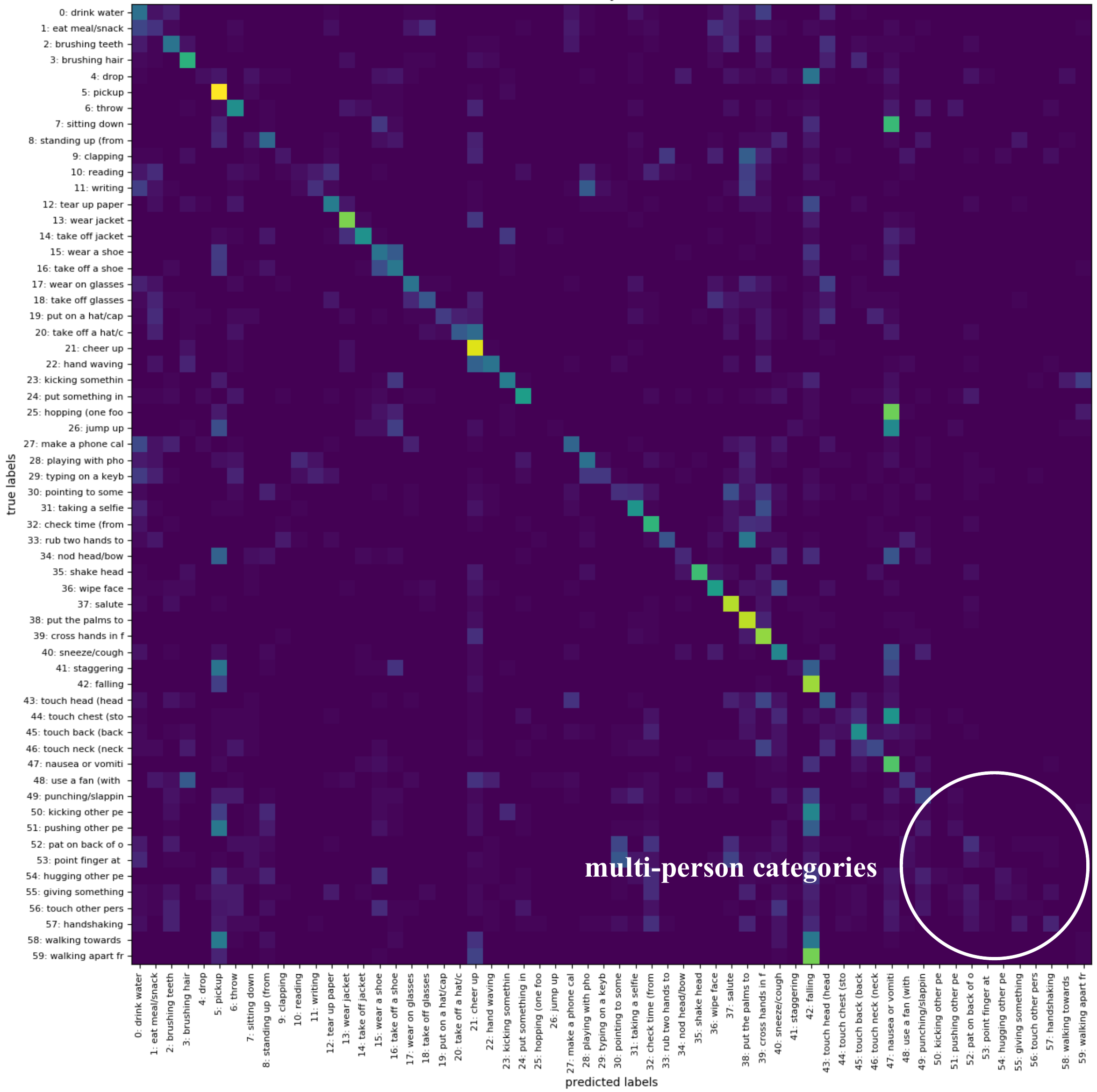}
    \captionof{figure}{
        \cpfont
        We render a version of the synthetic data with 10 sequences per action,
        where we only insert a single person per video.
        When trained with this data, the two-person interaction categories (last 11 classes)
        are mostly misclassified on the real NTU CVS 0$^\circ$ view test data
        The confusions suggest that it is important to model multi-person cases in the synthetic data.
    }
    \label{fig:app:confmatsingle}
\end{figure}

\vspace{0.1cm}
\noindent\textbf{Confusion matrices.}
We analyze two confusion matrices in Figure~\ref{fig:app:confmat}:
training only on the synthetic data
and (i)~testing on the synthetic test set;
and (ii)~testing on the real NTU CVS 0$^\circ$ view test set.
The confused classes are highlighted on the figure.
The confusions on both test sets suggest that
the fine-grained action classes require more precise body motions,
such as
\{\textit{clapping}, \textit{rub two hands together}\},
and
\{\textit{reading}, \textit{writing}\}. Other confusions
include object interaction categories
(e.g.~\{\textit{put on a hat}, \textit{brushing hair}\}
and
\{\textit{typing on a keyboard}, \textit{writing}\}), which can be explained
by the fact that synthetic data does not simulate objects.
These confusions are mostly resolved when 
training with both real and synthetic data.

\vspace{-0.3cm}
\subsection{Synthetic+Real training}
\label{subsec:app:synthreal}

\vspace{0.1cm}
\noindent\textbf{Amount of additional synthetic data.}
In
\if\sepappendix1{Figure~6 of the main paper,}
\else{Figure~\ref{fig:synthamount} of the main paper,}
\fi
we plotted the performance against the amount of
action sequences in the training set for both
synthetic and real datasets. Here,
we also report Synth+Real training performance
when the Real data is fixed and uses all the action
sequences available (i.e., 220 sequences per action),
and the Synth data is gradually increased.
Table~\ref{table:app:crosses} summarizes the results.
Increasing the amount of synthetic data improves
the performance. The improvement can be observed already
at the challenging $90^\circ$ view with as little
synthetic data as 10 sequences per action (57.2\% vs 53.6\%).
Using all the motions shows the most benefit as expected.

\vspace{0.1cm}
\noindent\textbf{Synth+Real training strategies.}
In all our experiments, we combine the training sets of
synthetic and real data to train jointly for both datasets,
which we referred as Synth+Real. Here, we investigate
whether a different strategy, such as using synthetic data
as pre-training (as in Varol~et~al.~\cite{varol2017surreal}),
would be more effective.
In Table~\ref{table:app:strategies},
we present several variations of training strategies.
We conclude that our Synth+Real, is simple yet effective,
while marginal gains can be obtained by continuing with
fine-tuning only on Real data.

\begin{table}[t]
    \centering
    \resizebox{0.79\linewidth}{!}{
        \begin{tabular}{lccc}
            \toprule
            & $0^\circ$  & $45^\circ$  & $90^\circ$ \\
            \midrule
            Real(220)              & 86.9 & 74.5 & 53.6 \\
            \midrule
            Synth(10) + Real(220)  & 85.5 & 74.7 & 57.2 \\
            Synth(30) + Real(220)  & 85.2 & 77.4 & 61.8 \\
            Synth(60) + Real(220)  & 87.6 & 78.7 & 62.2 \\
            Synth(100) + Real(220) & 87.7 & 78.8 & 63.7 \\
            Synth(220) + Real(220) & \textbf{89.1} & \textbf{82.0} & \textbf{67.1} \\
            \bottomrule
        \end{tabular}
    }
    \caption{
        \cpfont
        \textbf{Amount of synthetic data addition:}
        We experiment with Synth+Real RGB training while changing the number of sequences
        per action in the synthetic data and using all the real data. We conclude that
        using all available motion sequences improves the performance over taking a subset,
        confirming the importance of motion diversity. Results are reported on the NTU CVS protocol.}
    \label{table:app:crosses}
\end{table}

\begin{table}[t]
    \centering
    \resizebox{0.69\linewidth}{!}{
        \begin{tabular}{l ccc | c}
            \toprule
            & \multicolumn{3}{c|}{Real} & Synth \\
            Training          & $0^\circ$ & $45^\circ$ & $90^\circ$ & All \\
            \midrule
            S                 & 54.0 & 49.5 & 42.7 & 70.4 \\
            R                 & 86.9 & 74.5 & 53.6 & 18.1 \\
            S+R               & 89.1 & \textbf{82.0} & 67.1 & 71.0 \\
            \midrule
            S+R$\rightarrow$R & \textbf{89.9} & 81.9 & \textbf{67.5} & 73.1 \\
            S$\rightarrow$R   & 84.1 & 77.5 & 66.2 & 65.6 \\
            S$\rightarrow$S+R & 81.6 & 75.1 & 63.6 & \textbf{73.3} \\
            R$\rightarrow$S+R & 84.3 & 75.1 & 59.9 & 56.7 \\
            \bottomrule
        \end{tabular}
    }
    \caption{
        \cpfont
        \textbf{Training strategies with Synthetic+Real:}
        The bottom part of this table presents
        additional results for different training
        strategies on the NTU CVS protocol and the
        synthetic test set. The arrow A$\rightarrow$B
        denotes training first on A and then fine-tuning
        on B dataset. S and R stand for Synthetic and Real, respectively.
        Pre-training
        only on one dataset is suboptimal (last three rows).
        Our choice of training by mixing S+R from scratch
        is simple yet effective.
        Marginal gains can be obtained
        by continuing training only on Real data (S+R$\rightarrow$R).}
    \label{table:app:strategies}
    \vspace{-0.3cm}
\end{table}

\vspace{-0.3cm}
\subsection{Performance breakdown for object-related actions}
\label{subsec:app:ntuobjects}

While the NTU
dataset is mainly targeted for skeleton-based action recognition,
many actions involve object interactions. 
In Table~\ref{table:app:objectrelated},
we analyze the performance breakdown into
action categories with and without objects.
We notice that the object-related
actions have lower performance than body-only
counterparts even when trained with Real data.
The gap is higher when only synthetic training is used
since we simulate only humans, without objects.

\begin{table}
    \centering
    \resizebox{0.82\linewidth}{!}{
        \begin{tabular}{llccc}
            \toprule
            && $0^\circ$ & $45^\circ$ & $90^\circ$ \\
            \midrule
            \multirow{3}{*}{Real} & All actions & 86.9 & 74.5 & 53.6 \\
            & Human-body & 88.3 & 75.8 & 58.3 \\
            & Object-related & 85.2 & 73.0 & 48.1 \\
            \midrule
            \multirow{3}{*}{Synth} & All actions & 58.1 & 52.8 & 45.3 \\
            & Human-body & 62.9 & 60.5 & 55.9 \\
            & Object-related & 52.4 & 43.9 & 33.0 \\
            \midrule
            \multirow{3}{*}{Synth+Real} & All actions & 89.7 & 82.0 & 69.0 \\
            & Human-body & 90.3 & 83.2 & 71.2 \\
            & Object-related & 89.1 & 80.5 & 66.3 \\
            \bottomrule
        \end{tabular}
    }
    \caption{
        \cpfont
        \textbf{Object-related vs human-body actions:}
        We report the performance breakdown into 28 object-related
        and 32 human-body actions for the NTU CVS protocol. In all three setups,
        Real, Synth, and Synth+Real, human-body action categories
        have higher performance than object-related categories.
    }
    \label{table:app:objectrelated}
\end{table}

\subsection{Pretraining on Kinetics}
\label{subsec:app:pretraining}
Throughout the paper, the networks for NTU training are randomly initialized (i.e., scratch). Here, we investigate whether there is any gain from Kinetics~\cite{Kinetics} pretraining.
Table~\ref{table:app:pretraining} summarizes the results.
We refer to the table caption for the interpretation.

\begin{table}
    \centering
    \resizebox{0.99\linewidth}{!}{
        \begin{tabular}{llccc|ccc}
            \toprule
            & & \multicolumn{3}{c}{uniform} & \multicolumn{3}{c}{non-uniform}\\
            Optimizer & Pretraining & $0^\circ$ & $45^\circ$ & $90^\circ$ & $0^\circ$ & $45^\circ$ & $90^\circ$\\
            \midrule
            \multirow{3}{*}{RMSProp} & Kinetics [linear] & 42.5 & 33.3 & 27.2 & 48.1 & 37.7 & 31.1 \\
            & Kinetics [e2e] & 81.9 & 66.9 & 43.8 & 85.6 & 72.2 & 51.3 \\
            & Scratch & 
            83.9 & 67.9 & 42.9 & \textbf{86.9} & \textbf{74.5} & \textbf{53.6} \\
            \midrule
            \multirow{3}{*}{SGD} & Kinetics [linear] & 39.8 & 31.7 & 26.2 & 45.4 & 35.1 & 28.8 \\
            & Kinetics [e2e] & 89.4 & 69.8 & 40.5 & \textbf{91.9} & \textbf{78.9} & \textbf{55.4} \\
            & Scratch & 81.9 & 65.8 & 41.4 & 85.3 & 72.8 & 52.5 \\
            \bottomrule
        \end{tabular}
    }
    \caption{
        \cpfont
        \textbf{Effect of pretraining:} We measure the effect of 
        pretraining with Kinetics~\cite{Kinetics} over random initialization (real setting on the NTU CVS split).
        Interestingly, we do not observe
        improvements when the RMSProp optimizer is used, whereas SGD
        can improve the baselines from 53.6\% to 55.4\%. We note that
        this is still marginal compared to the boost we gain from synthetic data (69.0\% in Table~\ref{table:ntusynth}). Training only a linear layer on frozen features as opposed to end-to-end (e2e) finetuning
        is also suboptimal.
    }
    \label{table:app:pretraining}
\end{table}

\subsection{Non-uniform frame sampling}
\label{subsec:app:nonuniform}
In this section, we explore the proposed
frame sampling strategy further.

First, we confirm that the benefits of
non-uniform sampling applies also
to the flow stream. Since flow is estimated
online during training, we can
compute flow between any two frames.
Note that the flow estimation method
is learned on 2 consecutive frames, therefore
it produces noisy estimates for large displacements.
However, even with this noise,
in Table~\ref{table:app:framesamplingflow},
we demonstrate advantages of non-uniform
sampling over consecutive for the flow stream.

\begin{table}[t]
    \centering
    \resizebox{0.69\linewidth}{!}{
        \begin{tabular}{lccc}
            \toprule
            & $0^\circ$ & $45^\circ$ & $90^\circ$ \\
            \midrule
            Flow [uniform]     & 80.6 & 68.3 & 44.7 \\
            Flow [non-uniform] & \textbf{82.8} & \textbf{70.6} & \textbf{49.7} \\
            \bottomrule
        \end{tabular}
    }
    \vspace{-0.2cm}
    \caption{
        \cpfont
        \textbf{Frame sampling for the flow stream:}
        Training and testing on the real NTU CVS split. We confirm
        that the non-uniform sampling is beneficial also for the flow stream
        even though the flow estimates can be noisy between non-uniform frames.}
    \label{table:app:framesamplingflow}
\end{table}

\begin{table}[t]
    \centering
    \resizebox{0.99\linewidth}{!}{
        \begin{tabular}{lccc|ccc}
            \toprule
            & \multicolumn{6}{c}{Test mode} \\
            &  \multicolumn{3}{c|}{uniform}    &      \multicolumn{3}{c}{non-uniform} \\
            &$0^\circ$ & $45^\circ$  & $90^\circ$ &$0^\circ$ & $45^\circ$  & $90^\circ$  \\
            \midrule
            Train uniform   & \textbf{83.9} & \textbf{67.9} & \textbf{42.9} &    27.5&  20.6 & 13.8 \\
            Train non-uniform &     32.1  & 21.1 & 12.4 &    \textbf{86.9} &  \textbf{74.5} &  \textbf{53.6} \\
            \bottomrule
        \end{tabular}
    }
    \vspace{-0.2cm}
    \caption{
        \cpfont
        \textbf{Train/test modes:}
        Training and testing on the real NTU CVS split.
        The frame sampling mode should be the same at training and test times.}
    \label{table:app:framesampling}
\end{table}

\begin{table}%
    \centering
    \resizebox{0.99\linewidth}{!}{
        \begin{tabular}{lccc|ccc}
            \toprule
            & \multicolumn{6}{c}{Test mode} \\
            & \multicolumn{3}{c|}{ordered}    &  \multicolumn{3}{c}{non-ordered} \\
            &$0^\circ$ & $45^\circ$  & $90^\circ$ &$0^\circ$ & $45^\circ$  & $90^\circ$ \\
            \midrule
            Train ordered &    \textbf{86.9} &  \textbf{74.5} &  \textbf{53.6} &  16.4 & 13.1 &  8.1 \\
            Train non-ordered &    67.2 & 50.2 & 32.8 &  \textbf{72.7} &  \textbf{57.8} & \textbf{37.2} \\
            \bottomrule
        \end{tabular}
    }
    \caption{
        \cpfont
        \textbf{Frame order:}
        Training and testing on the real NTU CVS split. Preserving the order of frames in         
        non-uniform sampling is important. The confusion
        matrix in Figure~\ref{fig:app:confmatnonorder} shows that the mistakes are often among
        `symmetric' action classes such as \textit{sitting up} and \textit{standing up}. Order-aware
        models fail drastically when tested non-ordered, as expected.}
    \label{table:app:frameorder}
\end{table}

\begin{table}
    \centering
    \resizebox{0.99\linewidth}{!}{
        \begin{tabular}{lccc}
            \toprule
            & $0^\circ$ & $45^\circ$ & $90^\circ$ \\
            \midrule
            1. Uniform, random shift, original fps & 83.9 & 67.9 & 42.9 \\
            2. Uniform, random shift, random fps & 84.1 & 69.9 & 48.9 \\
            \midrule
            3. Uniform, random shift, with smallest fps & 85.6 & 74.0 & 54.0 \\
            4. Hybrid (random within uniform segments) & 85.3 & 73.9 & \textbf{54.3} \\
            5. Non-uniform random & \textbf{86.9} & \textbf{74.5} & 53.6 \\
            \bottomrule
        \end{tabular}
    }
    \caption{
        \cpfont
        \textbf{Frame sampling alternatives:}
        Training and testing on the real NTU CVS split.
        We explore alternative random sampling schemes besides
        uniform baseline (1) and the random non-uniform (5).
        (2) applies a random frame rate similar to~\cite{Zhu2018RandomTS}.
        However, (3) with a fixed frame rate of maximum temporal skip
        outperforms the random fps, suggesting the importance of long-term
        span. (4) the hybrid approach similar to \cite{TSN2016ECCV,ECO_eccv18}
        increases the data augmentation while ensuring long-term context.
        (5) our fully random sampling maximizes the data augmentation.
        The approaches (3)(4)(5) perform similarly outperforming (1)(2).
    }
    \label{table:app:framesamplingfull}
\end{table}

Next, we present our experiments about
the testing modes as mentioned in
\if\sepappendix1{Section~3.3 of the main paper.}
\else{Section~\ref{subsec:approach:training} of the main paper.}
\fi
Table~\ref{table:app:framesampling}
suggests that the training and testing modes
should be the same for both uniform and
non-uniform samplings. The convolutional
filters adapt to certain training statistics,
which should be preserved at test time.

\begin{figure}
    \centering
    {\scriptsize Train/Test: Real 0$^\circ$ non-uniform, \textbf{non-ordered}} \\
    \includegraphics[width=0.99\linewidth]{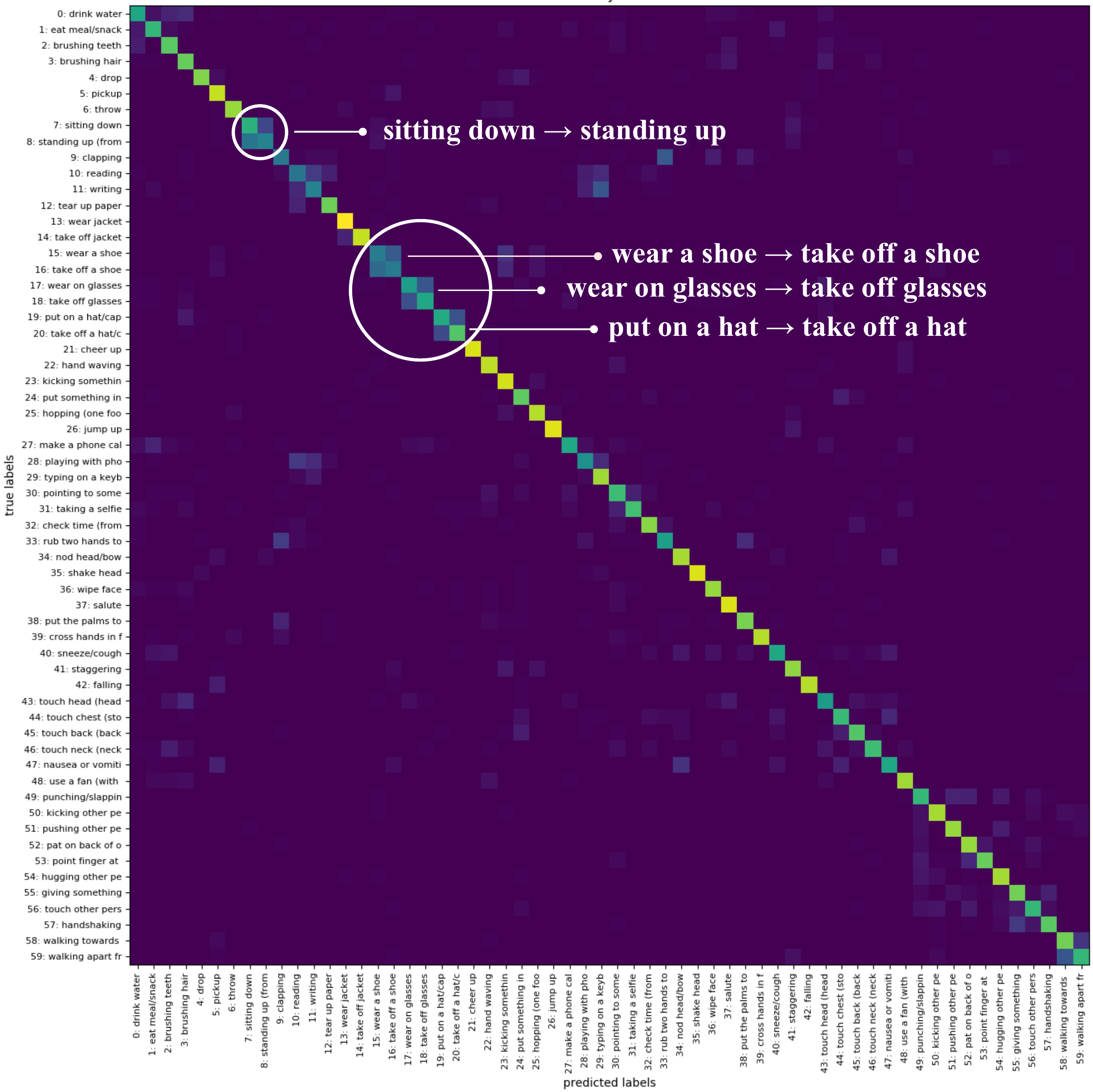}
    \captionof{figure}{
        \cpfont
        We present the confusion matrix for the non-ordered training
        explained in Table~\ref{table:app:frameorder}. The classes that require
        the temporal order to be distinguished are confused as expected.
        The training and test is performed on the real NTU CVS 0$^\circ$ view split.
    }
    \label{fig:app:confmatnonorder}
\end{figure}

We then investigate the importance
of the frame order when we randomly
sample non-uniformly. We preserve the
temporal order in all our experiments,
except in Table~\ref{table:app:frameorder},
where we experiment with a shuffled order.
In this case, we observe a significant performance
drop which can be explained by the confusion matrix
in Figure~\ref{fig:app:confmatnonorder}.
The action classes such as \textit{wearing} and \textit{taking off}
are heavily confused when the order is not
preserved.
This experiment allows detecting action categories
that are temporally symmetric~\cite{price2019_RetroActionsLearning}.
We also observe that the ordered
model fails when tested in non-ordered mode,
which indicates that the convolutional kernels
become highly order-aware.

Finally, we experiment with other frame sampling
alternatives in Table~\ref{table:app:framesamplingfull}.
See the table caption for interpretation of the results.


%% file: main.bbl
\begin{thebibliography}{10}
\providecommand{\url}[1]{\texttt{#1}}
\providecommand{\urlprefix}{URL }
\providecommand{\doi}[1]{https://doi.org/#1}

\bibitem{cmu_mocap}
{Carnegie-Mellon Mocap Database}. \url{http://mocap.cs.cmu.edu/}

\bibitem{Badler1993}
Badler, N.I., Phillips, C.B., Webber, B.L.: Simulating Humans: Computer
  Graphics Animation and Control. Oxford University Press, Inc., New York, NY,
  USA (1993)

\bibitem{Baradel17}
Baradel, F., Wolf, C., Mille, J.: Pose-conditioned spatio-temporal attention
  for human action recognition. CoRR  \textbf{abs/1703.10106} (2017)

\bibitem{baradel18}
Baradel, F., Wolf, C., Mille, J., Taylor, G.W.: Glimpse clouds: Human activity
  recognition from unstructured feature points. In: CVPR (2018)

\bibitem{Bogo2016smplify}
Bogo, F., Kanazawa, A., Lassner, C., Gehler, P., Romero, J., Black, M.J.: Keep
  it {SMPL}: Automatic estimation of {3D} human pose and shape from a single
  image. In: ECCV (2016)

\bibitem{Carreira2017}
Carreira, J., Zisserman, A.: Quo vadis, action recognition? {A} new model and
  the {Kinetics} dataset. In: CVPR (2017)

\bibitem{chen2020deep}
Chen, C.F., Panda, R., Ramakrishnan, K., Feris, R., Cohn, J., Oliva, A., Fan,
  Q.: Deep analysis of {CNN}-based spatio-temporal representations for action
  recognition. arXiv preprint arXiv:2010.11757  (2020)

\bibitem{synthetic_cohenor}
Chen, W., Wang, H., Li, Y., Su, H., Wang, Z., Tu, C., Lischinski, D., Cohen-Or,
  D., Chen, B.: Synthesizing training images for boosting human {3D} pose
  estimation. In: 3DV (2016)

\bibitem{crasto2019mars}
Crasto, N., Weinzaepfel, P., Alahari, K., Schmid, C.: {MARS}:
  {M}otion-augmented {RGB} stream for action recognition. In: CVPR (2019)

\bibitem{DeSouza:Procedural:CVPR2017}
De~Souza, C.R., Gaidon, A., Cabon, Y., L{\'{o}}pez~Pe{\~{n}}a, A.M.: Procedural
  generation of videos to train deep action recognition networks. In: CVPR
  (2017)

\bibitem{Doersch2019}
Doersch, C., Zisserman, A.: Sim2real transfer learning for {3D} pose
  estimation: motion to the rescue. CoRR  \textbf{abs/1907.02499} (2019)

\bibitem{dosovitskiy_flownet}
Dosovitskiy, A., Fischer, P., Ilg, E., Hausser, P., Hazirbas, C., Golkov, V.,
  van~der Smagt, P., Cremers, D., Brox, T.: {FlowNet}: Learning optical flow
  with convolutional networks. In: ICCV (2015)

\bibitem{fang2017rmpe}
Fang, H.S., Xie, S., Tai, Y.W., Lu, C.: {RMPE}: Regional multi-person pose
  estimation. In: ICCV (2017)

\bibitem{Farhadi2008}
Farhadi, A., Tabrizi, M.K.: Learning to recognize activities from the wrong
  view point. In: ECCV (2008)

\bibitem{feichtenhofer2019slowfast}
Feichtenhofer, C., Fan, H., Malik, J., He, K.: {SlowFast} networks for video
  recognition. In: ICCV (2019)

\bibitem{Feichtenhofer16}
Feichtenhofer, C., Pinz, A., Zisserman, A.: Convolutional two-stream network
  fusion for video action recognition. In: CVPR (2016)

\bibitem{Deep3DPose}
Ghezelghieh, M.F., Kasturi, R., Sarkar, S.: Learning camera viewpoint using
  {CNN} to improve {3D} body pose estimation. In: 3DV (2016)

\bibitem{HaraCVPR2018}
Hara, K., Kataoka, H., Satoh, Y.: Can spatiotemporal {3D} {CNNs} retrace the
  history of {2D} {CNNs} and {ImageNet}? In: CVPR (2018)

\bibitem{hasson19_obman}
Hasson, Y., Varol, G., Tzionas, D., Kalevatykh, I., Black, M.J., Laptev, I.,
  Schmid, C.: Learning joint reconstruction of hands and manipulated objects.
  In: CVPR (2019)

\bibitem{He2016}
He, K., Zhang, X., Ren, S., Sun, J.: Deep residual learning for image
  recognition. In: CVPR (2015)

\bibitem{LSTM1997}
Hochreiter, S., Schmidhuber, J.: Long short-term memory. Neural Computation
  \textbf{9}(8),  1735--1780 (1997)

\bibitem{Hoffmann:GCPR:2019}
Hoffmann, D.T., Tzionas, D., Black, M.J., Tang, S.: Learning to train with
  synthetic humans. In: GCPR (2019)

\bibitem{joule2017}
Hu, J.F., Zheng, W.S., Lai, J., Jianguo, Z.: Jointly learning heterogeneous
  features for {RGB-D} activity recognition. IEEE Transactions on Pattern
  Analysis and Machine Intelligence  \textbf{39}(11),  2186--2200 (2017)

\bibitem{hri40}
Ji, Y., Xu, F., Yang, Y., Shen, F., Shen, H.T., Zheng, W.: A large-scale
  {RGB-D} database for arbitrary-view human action recognition. In: ACMMM
  (2018)

\bibitem{Jingtian2018}
Jingtian, Z., Shum, H., Han, J., Shao, L.: Action recognition from arbitrary
  views using transferable dictionary learning. IEEE Transactions on Image
  Processing  (2018)

\bibitem{junejo2011}
Junejo, I.N., Dexter, E., Laptev, I., Perez, P.: View-independent action
  recognition from temporal self-similarities. IEEE Transactions on Pattern
  Analysis and Machine Intelligence  \textbf{33}(1),  172--185 (2011)

\bibitem{hmrKanazawa17}
Kanazawa, A., Black, M.J., Jacobs, D.W., Malik, J.: End-to-end recovery of
  human shape and pose. In: CVPR (2018)

\bibitem{humanMotionKZFM19}
Kanazawa, A., Zhang, J.Y., Felsen, P., Malik, J.: Learning {3D} human dynamics
  from video. In: CVPR (2019)

\bibitem{Kinetics}
Kay, W., Carreira, J., Simonyan, K., Zhang, B., Hillier, C., Vijayanarasimhan,
  S., Viola, F., Green, T., Back, T., Natsev, P., Suleyman, M., Zisserman, A.:
  The {Kinetics} human action video dataset. CoRR  \textbf{abs/1705.06950}
  (2017)

\bibitem{KeCVPR17}
Ke, Q., Bennamoun, M., An, S., Sohel, F., Boussaid, F.: A new representation of
  skeleton sequences for {3D} action recognition. In: CVPR (2017)

\bibitem{VIBECVPR2020}
Kocabas, M., Athanasiou, N., Black, M.J.: {VIBE}: Video inference for human
  body pose and shape estimation. In: CVPR (2020)

\bibitem{kolotouros2019spin}
Kolotouros, N., Pavlakos, G., Black, M.J., Daniilidis, K.: Learning to
  reconstruct {3D} human pose and shape via model-fitting in the loop. In: ICCV
  (2019)

\bibitem{kolotouros2019cmr}
Kolotouros, N., Pavlakos, G., Daniilidis, K.: Convolutional mesh regression for
  single-image human shape reconstruction. In: CVPR (2019)

\bibitem{KongTIP2017}
Kong, Y., Ding, Z., Li, J., Fu, Y.: Deeply learned view-invariant features for
  cross-view action recognition. IEEE Transactions on Image Processing
  \textbf{26}(6),  3028--3037 (2017)

\bibitem{Kong2018HumanAR}
Kong, Y., Fu, Y.: Human action recognition and prediction: {A} survey. CoRR
  \textbf{abs/1806.11230} (2018)

\bibitem{Kortylewski2018}
Kortylewski, A., Egger, B., Schneider, A., Gerig, T., Morel-Forster, A.,
  Vetter, T.: Empirically analyzing the effect of dataset biases on deep face
  recognition systems. In: CVPRW (2018)

\bibitem{lassner2017up}
Lassner, C., Romero, J., Kiefel, M., Bogo, F., Black, M.J., Gehler, P.V.: Unite
  the people: Closing the loop between {3D} and {2D} human representations. In:
  CVPR (2017)

\bibitem{LeCun1989cnn}
LeCun, Y., Boser, B.E., Denker, J.S., Henderson, D., Howard, R.E., Hubbard,
  W.E., Jackel, L.D.: Backpropagation applied to handwritten zip code
  recognition. Neural Computation  \textbf{1}(4),  541--551 (1989)

\bibitem{NIPS2018_7401}
Li, J., Wong, Y., Zhao, Q., Kankanhalli, M.: Unsupervised learning of
  view-invariant action representations. In: NeurIPS (2018)

\bibitem{Li2018}
Li, W., Xu, Z., Xu, D., Dai, D., Gool, L.V.: Domain generalization and
  adaptation using low rank exemplar {SVM}s. IEEE Transactions on Pattern
  Analysis and Machine Intelligence  \textbf{40}(5),  1114--1127 (2018)

\bibitem{lin2019tsm}
Lin, J., Gan, C., Han, S.: {TSM}: {T}emporal shift module for efficient video
  understanding. In: ICCV (2019)

\bibitem{Liu2019Temp}
Liu, J., Akhtar, N., Mian, A.: Temporally coherent full {3D} mesh human pose
  recovery from monocular video. CoRR  \textbf{abs/1906.00161} (2019)

\bibitem{Liu2019}
Liu, J., Rahmani, H., Akhtar, N., Mian, A.: Learning human pose models from
  synthesized data for robust {RGB-D} action recognition. International Journal
  of Computer Vision (IJCV)  (2019)

\bibitem{Liu2011CrossviewAR}
Liu, J., Shah, M., Kuipers, B., Savarese, S.: Cross-view action recognition via
  view knowledge transfer. In: CVPR (2011)

\bibitem{LiuECCV2016}
Liu, J., Shahroudy, A., Xu, D., Wang, G.: Spatio-temporal {LSTM} with trust
  gates for {3D} human action recognition. In: ECCV (2016)

\bibitem{LiuCVPR2017}
Liu, J., Wang, G., Hu, P., Duan, L.Y., Kot, A.C.: Global context-aware
  attention {LSTM} networks for {3D} action recognition. In: CVPR (2017)

\bibitem{Liu2017ESV}
Liu, M., Liu, H., Chen, C.: Enhanced skeleton visualization for view invariant
  human action recognition. Pattern Recognition  \textbf{68}(C),  346--362
  (2017)

\bibitem{Liu_2018_CVPR}
Liu, M., Yuan, J.: Recognizing human actions as the evolution of pose
  estimation maps. In: CVPR (2018)

\bibitem{Liu2019_viewinvariant}
Liu, Y., Lu, Z., Li, J., Yang, T.: Hierarchically learned view-invariant
  representations for cross-view action recognition. IEEE Transactions on
  Circuits and Systems for Video Technology  (2019)

\bibitem{smpl2015}
Loper, M., Mahmood, N., Romero, J., Pons-Moll, G., Black, M.J.: {SMPL}: A
  skinned multi-person linear model. In: SIGGRAPH Asia (2015)

\bibitem{luo_eccv18_graph}
Luo, Z., Hsieh, J.T., Jiang, L., Niebles, J.C., Fei-Fei, L.: Graph distillation
  for action detection with privileged information. In: ECCV (2018)

\bibitem{Luvizon20182D3DPE}
Luvizon, D.C., Picard, D., Tabia, H.: {2D/3D} pose estimation and action
  recognition using multitask deep learning. In: CVPR (2018)

\bibitem{Lv2007}
Lv, F., Nevatia, R.: Single view human action recognition using key pose
  matching and viterbi path searching. In: CVPR (2007)

\bibitem{AMASS:ICCV:2019}
Mahmood, N., Ghorbani, N., Troje, N.F., Pons-Moll, G., Black, M.J.: {AMASS}:
  Archive of motion capture as surface shapes. In: ICCV (2019)

\bibitem{MarinVGL10}
Marin, J., Vazquez, D., Geronimo, D., Lopez, A.M.: Learning appearance in
  virtual scenarios for pedestrian detection. In: CVPR (2010)

\bibitem{Masi2019Face}
Masi, I., Tran, A.T., Hassner, T., Sahin, G., Medioni, G.: Face-specific data
  augmentation for unconstrained face recognition. International Journal of
  Computer Vision (IJCV)  (2019)

\bibitem{newell2016hourglass}
Newell, A., Yang, K., Deng, J.: Stacked hourglass networks for human pose
  estimation. In: ECCV (2016)

\bibitem{omran2018nbf}
Omran, M., Lassner, C., Pons-Moll, G., Gehler, P.V., Schiele, B.: Neural body
  fitting: Unifying deep learning and model-based human pose and shape
  estimation. In: 3DV (2018)

\bibitem{pavlakos2018humanshape}
Pavlakos, G., Zhu, L., Zhou, X., Daniilidis, K.: Learning to estimate 3{D}
  human pose and shape from a single color image. In: CVPR (2018)

\bibitem{Pishchulin2012}
Pishchulin, L., Jain, A., Andriluka, M., Thorm{\"a}hlen, T., Schiele, B.:
  Articulated people detection and pose estimation: Reshaping the future. In:
  CVPR (2012)

\bibitem{price2019_RetroActionsLearning}
Price, W., Damen, D.: {Retro-Actions}: Learning 'close' by time-reversing
  'open' videos  (2019)

\bibitem{virtualhome2018}
Puig, X., Ra, K., Boben, M., Li, J., Wang, T., Fidler, S., Torralba, A.:
  {VirtualHome}: Simulating household activities via programs. In: CVPR (2018)

\bibitem{Qian_2018_ECCV}
Qian, X., Fu, Y., Xiang, T., Wang, W., Qiu, J., Wu, Y., Jiang, Y.G., Xue, X.:
  Pose-normalized image generation for person re-identification. In: ECCV
  (2018)

\bibitem{Rahmani2014}
Rahmani, H., Mahmood, A., Huynh, D., Mian, A.: Histogram of oriented principal
  components for cross-view action recognition. IEEE Transactions on Pattern
  Analysis and Machine Intelligence  \textbf{38}(12),  2430--2443 (2016)

\bibitem{NKTM}
Rahmani, H., Mian, A.: Learning a non-linear knowledge transfer model for
  cross-view action recognition. In: CVPR (2015)

\bibitem{Rahmani2016Novel}
Rahmani, H., Mian, A.: {3D} action recognition from novel viewpoints. In: CVPR
  (2016)

\bibitem{rahmani2018}
Rahmani, H., Mian, A., Shah, M.: Learning a deep model for human action
  recognition from novel viewpoints. IEEE Transactions on Pattern Analysis and
  Machine Intelligence  \textbf{40}(3),  667--681 (2018)

\bibitem{Sakoe1978DynamicPA}
Sakoe, H., Chiba, S.: Dynamic programming algorithm optimization for spoken
  word recognition. IEEE Transactions on Acoustics, Speech, and Signal
  Processing  \textbf{26}(1),  43--49 (1978)

\bibitem{NTURGBD}
Shahroudy, A., Liu, J., Ng, T.T., Wang, G.: {NTU RGB+D}: A large scale dataset
  for {3D} human activity analysis. In: CVPR (2016)

\bibitem{Shi19dg}
{Shi}, L., {Zhang}, Y., {Cheng}, J., {Lu}, H.: Skeleton-based action
  recognition with directed graph neural networks. In: CVPR (2019)

\bibitem{Shi19twostream}
{Shi}, L., {Zhang}, Y., {Cheng}, J., {Lu}, H.: Two-stream adaptive graph
  convolutional networks for skeleton-based action recognition. In: CVPR (2019)

\bibitem{shotton2011}
Shotton, J., Fitzgibbon, A., Cook, M., Sharp, T., Finocchio, M., Moore, R.,
  Kipman, A., Blake, A.: Real-time human pose recognition in parts from single
  depth images. In: CVPR (2011)

\bibitem{Si19}
{Si}, C., {Chen}, W., {Wang}, W., {Wang}, L., {Tan}, T.: An attention enhanced
  graph convolutional {LSTM} network for skeleton-based action recognition. In:
  CVPR (2019)

\bibitem{simonyan_twostream}
Simonyan, K., Zisserman, A.: Two-stream convolutional networks for action
  recognition in videos. In: NeurIPS (2014)

\bibitem{UCF101}
Soomro, K., Roshan~Zamir, A., Shah, M.: {UCF101}: A dataset of 101 human
  actions classes from videos in the wild. CRCV-TR-12-01  (2012)

\bibitem{Su_2015_ICCV}
Su, H., Qi, C.R., Li, Y., Guibas, L.J.: Render for {CNN}: Viewpoint estimation
  in images using {CNN}s trained with rendered {3D} model views. In: ICCV
  (2015)

\bibitem{surreactpage}
{SURREACT} project page. \url{https://www.di.ens.fr/willow/research/surreact/}

\bibitem{Tieleman2012}
Tieleman, T., Hinton, G.: Lecture 6.5---{RMSprop}: Divide the gradient by a
  running average of its recent magnitude. COURSERA: Neural Networks for
  Machine Learning (2012)

\bibitem{C3D}
Tran, D., Bourdev, L., Fergus, R., Torresani, L., Paluri, M.: Learning
  spatiotemporal features with {3D} convolutional networks. In: ICCV (2015)

\bibitem{tung2017selfsupervised}
Tung, H.Y.F., Tung, H.W., Yumer, E., Fragkiadaki, K.: Self-supervised learning
  of motion capture. In: NeurIPS (2017)

\bibitem{varol18_bodynet}
Varol, G., Ceylan, D., Russell, B., Yang, J., Yumer, E., Laptev, I., Schmid,
  C.: {BodyNet}: Volumetric inference of {3D} human body shapes. In: ECCV
  (2018)

\bibitem{varol18_ltc}
Varol, G., Laptev, I., Schmid, C.: Long-term temporal convolutions for action
  recognition. IEEE Transactions on Pattern Analysis and Machine Intelligence
  \textbf{40}(6),  1510--1517 (2018)

\bibitem{varol2017surreal}
Varol, G., Romero, J., Martin, X., Mahmood, N., Black, M.J., Laptev, I.,
  Schmid, C.: Learning from synthetic humans. In: CVPR (2017)

\bibitem{Wang_2018_ECCV}
Wang, D., Ouyang, W., Li, W., Xu, D.: Dividing and aggregating network for
  multi-view action recognition. In: ECCV (2018)

\bibitem{NUCLA2014}
Wang, J., Nie, X., Xia, Y., Wu, Y., Zhu, S.C.: Cross-view action modeling,
  learning, and recognition. In: CVPR (2014)

\bibitem{TSN2016ECCV}
Wang, L., Xiong, Y., Wang, Z., Qiao, Y., Lin, D., Tang, X., {Van Gool}, L.:
  Temporal segment networks: Towards good practices for deep action
  recognition. In: ECCV (2016)

\bibitem{IXMAS2007}
Weinland, D., Boyer, E., Ronfard, R.: Action recognition from arbitrary views
  using {3D} exemplars. In: ICCV (2007)

\bibitem{Xie2017RethinkingSF}
Xie, S., Sun, C., Huang, J., Tu, Z., Murphy, K.: Rethinking spatiotemporal
  feature learning: Speed-accuracy trade-offs in video classification. In: ECCV
  (2017)

\bibitem{stgcn}
Yan, S., Xiong, Y., Lin, D.: Spatial temporal graph convolutional networks for
  skeleton-based action recognition. In: AAAI (2018)

\bibitem{yu_lsun}
Yu, F., Zhang, Y., Song, S., Seff, A., Xiao, J.: {LSUN}: Construction of a
  large-scale image dataset using deep learning with humans in the loop. CoRR
  \textbf{abs/1506.03365} (2015)

\bibitem{Yuille2018DeepNW}
Yuille, A.L., Liu, C.: Deep nets: What have they ever done for vision? CoRR
  \textbf{abs/1805.04025} (2018)

\bibitem{Zhang_2018_CVPR}
Zhang, D., Guo, G., Huang, D., Han, J.: {PoseFlow}: {A} deep motion
  representation for understanding human behaviors in videos. In: CVPR (2018)

\bibitem{Zhang2017ViewAdaptive}
Zhang, P., Lan, C., Xing, J., Zeng, W., Xue, J., Zheng, N.: View adaptive
  recurrent neural networks for high performance human action recognition from
  skeleton data. In: ICCV (2017)

\bibitem{Zheng2013}
Zheng, J., Jiang, Z.: Learning view-invariant sparse representations for
  cross-view action recognition. In: ICCV (2013)

\bibitem{Zheng2016}
Zheng, J., Jiang, Z., Chellappa, R.: Cross-view action recognition via
  transferable dictionary learning. IEEE Transactions on Image Processing
  \textbf{25}(6),  2542--2556 (2016)

\bibitem{Zhu2018RandomTS}
Zhu, Y., Newsam, S.: Random temporal skipping for multirate video analysis. In:
  ACCV (2018)

\bibitem{zb2017hand}
Zimmermann, C., Brox, T.: Learning to estimate {3D} hand pose from single {RGB}
  images. In: ICCV (2017)

\bibitem{ZolfaghariOSB17}
Zolfaghari, M., Oliveira, G.L., Sedaghat, N., Brox, T.: Chained multi-stream
  networks exploiting pose, motion, and appearance for action classification
  and detection. In: ICCV (2017)

\bibitem{ECO_eccv18}
Zolfaghari, M., Singh, K., Brox, T.: {ECO:} efficient convolutional network for
  online video understanding. In: ECCV (2018)

\end{thebibliography}
